%% file: main.tex
\documentclass{article}

\PassOptionsToPackage{numbers,compress}{natbib}
\usepackage[preprint]{neurips}

\usepackage[utf8]{inputenc}
\usepackage[T1]{fontenc}

\newcommand{\ourmodel}{\textsc{V2M-Zero} }
\newcommand{\ourmodela}{\textsc{V2M-Zero}}
\input{setup/package}

\input{setup/macros}

\usepackage{hyperref}
\usepackage{url}

\title{\textsc{V2M-Zero}: Zero-Pair Time-Aligned Video-to-Music Generation}

\author{
Yan-Bo Lin$^{1}$\thanks{Work done during an internship at Adobe Research.}\hspace{1em}
Jonah Casebeer$^{2}$\hspace{1em}
Long Mai$^{2}$\hspace{1em}
Aniruddha Mahapatra$^{2}$\\
\bfseries Gedas Bertasius$^{1}$\hspace{1em}
Nicholas J. Bryan$^{2}$
\\[0.5em]
\mdseries $^{1}$UNC Chapel Hill\hspace{1em}
$^{2}$Adobe Research
}

\begin{document}
\maketitle

\input{0_Abstract}

\input{1_intro}
\input{2_related_work}
\input{3_Method}

\input{4_result}
\input{5_Conclusion}

\begin{ack}
This work was supported by Laboratory for Analytic Sciences via NC State University, ONR Award N00014-23-1-2356, NIH Award R01HD11107402, and Sony Focused Research award.
\end{ack}

\newcount\cvprrulercount
\appendix

\input{6_appendix}

\clearpage

{
\bibliographystyle{ieee_fullname}
\bibliography{egbib}
}

\end{document}

%% file: setup/package.tex
\usepackage{comment}
\usepackage{tablefootnote}
\usepackage{paralist}
\usepackage{graphicx}
\usepackage{subcaption}
\usepackage{float}

\usepackage{algorithm}
\usepackage{algpseudocode}

\usepackage{booktabs}
\usepackage{multirow}
\usepackage{array}
\usepackage{enumitem}

\usepackage{graphicx}
\usepackage{mathtools}
\usepackage{amssymb,amsmath}

\usepackage{pifont}
\usepackage{xspace}

\usepackage{diagbox}
\usepackage{xcolor}
\usepackage[export]{adjustbox}
\usepackage{sidecap}

\usepackage{multirow, multicol, makecell, booktabs}
\definecolor{citecolor}{HTML}{0071bc}
\definecolor{linkcolor}{HTML}{ED1C24}

\usepackage[pagebackref,breaklinks,colorlinks,citecolor=blue]{hyperref}

\usepackage{wrapfig}
\usepackage{listings}

\definecolor{codegreen}{rgb}{0,0.6,0}
\definecolor{codegray}{rgb}{0.5,0.5,0.5}

\lstdefinestyle{pythonstyle}{
    commentstyle=\color{codegreen},
    keywordstyle=\color{blue},
    numberstyle=\tiny\color{codegray},
    stringstyle=\color{orange!80!black},
    basicstyle=\ttfamily\footnotesize,
    breaklines=true,
    numbers=left,
    numbersep=5pt,
    showtabs=false,
    showspaces=false,
    showstringspaces=false,
    frame=lines,
    backgroundcolor=\color{white},
}

%% file: setup/macros.tex
\def\eg{e.g.,~}
\def\ie{i.e.,~}

\newcommand{\figref}[1]{Figure~\ref{fig:#1}}
\newcommand{\tabref}[1]{Table~\ref{tab:#1}}

\long\def\ignorethis#1{}

\newlength\paramargin
\newlength\figmargin
\newlength\subfigmargin
\newlength\secmargin
\newlength\subsecmargin
\newlength\tabmargin
\newlength\eqmargin

\newcolumntype{C}[1]{>{\centering\let\newline\\\arraybackslash\hspace{0pt}}m{#1}}

\newcommand*\colourmark[1]{
  \expandafter\newcommand\csname #1xmark\endcsname{\textcolor{#1}{\ding{56}}}
}

\colourmark{blue}
\colourmark{red}

\newcommand*\colourchecksnow[1]{
  \expandafter\newcommand\csname #1snow\endcsname{\textcolor{#1}{\ding{100}}}
}

\colourchecksnow{cyan}

\definecolor{darkspringgreen}{rgb}{0, 0.6, 0.3}
\newcommand*\colourcheck[1]{
  \expandafter\newcommand\csname #1check\endcsname{\textcolor{#1}{\ding{52}}}
}

\colourcheck{blue}
\colourcheck{cadmiumgreen}

\newcommand*\colourtri[1]{
  \expandafter\newcommand\csname #1tri\endcsname{\textcolor{#1}{\ding{115}}}
}

\colourtri{black}

\makeatletter
\@namedef{ver@everyshi.sty}{}
\makeatother
\usepackage{tikzsymbols}
\newcommand*\colourcheckfire[1]{
  \expandafter\newcommand\csname #1fire\endcsname{\textcolor{#1}{\Fire}}
}
\colourcheckfire{red}

%% file: 0_Abstract.tex
\begin{abstract}
Generating music that temporally aligns with video events is challenging for existing text-to-music models, which lack fine-grained temporal control.
We introduce \ourmodela, a video-to-music generation approach that generates time-aligned music with disentangled time synchronization and semantic control (e.g. genre, mood) from video while requiring zero video-music pairs at training time.
Our method is motivated by a key observation: temporal synchronization requires matching when and how much change occurs, not what changes.
While musical and visual events differ semantically, they exhibit shared temporal structure that can be captured independently within each modality.
We capture this structure through event curves computed from intra-modal similarity using pretrained music and video encoders.
By measuring temporal change within each modality independently, these curves provide comparable representations across modalities.
This enables a simple training strategy: fine-tune a text-to-music model on music-event curves, then substitute video-event curves at inference without cross-modal training or paired data.
Across {OES-Pub}, {MovieGenBench-Music}, and {AIST++}, \ourmodel achieves state-of-the-art performance \emph{without any paired music-video data}, surpassing the strongest prior baselines per metric with {5--9\%} higher audio quality, {13--15\%} better semantic alignment, {21--52\%} improved temporal synchronization, and {28\%} higher beat alignment on dance videos.
We find similar results via a large crowd-source subjective listening test.
Our results validate that temporal alignment through within-modality features is not only effective for video-to-music generation but leads to \textit{better} performance over paired cross-modal supervision. Furthermore, our approach enables independent controls for timing and music style (e.g. genre, mood) for more controllable generation.
Results are available at~\url{https://genjib.github.io/v2m_zero/}
\end{abstract}

%% file: 1_intro.tex
\section{Introduction}\label{sec:intro}
\vspace{\secmargin}

Generative music is growing in popularity among creators from online influencers on social media platforms (\eg YouTube, Instagram, TikTok) to professionals in film, gaming, and advertising.
Such content creators seek music that both complements their video content and supports separate control over style and time synchronization.
While recent text-to-music (T2M) methods~\cite{arxiv25_yue,arxiv23_musiclm,musicgen,nips23_melody,arxiv23_Mousais,iclr24_MAGNET,tmlr25_t2m,arxiv25_ace_step,icml25_song_gen,arxiv24_flux_music,icml25_ditto} enable automatic music generation from textual prompts, they operate solely on text inputs and typically are not designed to generate time-synchronized music that matches a target video.
As a result, creators have to manually and tediously edit videos to fit the generated music for synchronization, a laborious and time-consuming process.
For instance, in a short product promotion video, musical cues must align with each reveal or motion highlight to create a strong and memorable impression within seconds.

The lack of time-synchronized text-to-music generation motivates video-to-music (V2M) models~\cite{lin2025vmas,audiox,cvpr25_filmcomposer,cvpr25_vidmuse,arxiv25_extending,ismir25_OES,aaai25_gvmgen,acmmm25_controllable,arxiv24_vidmusician,arxiv24_muvi,cvpr24_melfusion}, or the task of creating background music that is both temporally and semantically aligned with a given video.
Existing V2M methods typically rely on paired video–music datasets curated from online videos, entangling these two aspects of control.
Recent work~\cite{icassp25_sonique, vibe} explores an alternative direction by using pretrained multimodal large language models (MLLMs) to infer music prompts for video, which are then used as input to a pretrained T2M model. While promising for semantic alignment, such methods do not explicitly model temporal correspondence between visual and musical events, limiting user control. Furthermore, we note that it is now common to require licensed data for training music models~\cite{yinghao2024foundation} and licensing \textit{paired} music and video is significantly more complicated than licensing music and silent video separately~\cite{menell2022musiclicensing, rosario2024sync}.
This motivates our goal: time-synchronized video-to-music generation with indepenent timing and semantic controls, trained without paired video-music data.

We realize this with \ourmodela -- a video-to-music generation method that generates \textit{\textbf{time-synchronized}} music from videos with separate timing and semantic controls -- all \textit{\textbf{without}} paired video–music data (zero-pair).
Our key observation is that synchronization primarily depends on \textit{\textbf{when}} change occurs, rather than \textit{\textbf{what}} changes.
In practice, video and music synchronization often corresponds to (sparse) moments of interest or events over time (\eg video events of dancing and scene transitions match music events like beats, instrumental and/or dynamic changes).
Concretely, we represent events over time as event \textit{\textbf{curves}} computed from intra-modal similarity in encoder feature spaces. The result is a representation that captures perceptually relevant change across video and music~( \figref{dynamic}).
Crucially, our design decouples temporal structure from semantic and emotional grounding: event curves specify \textit{\textbf{when}} musical change should occur, while textual conditioning determines \textit{\textbf{how}} it should sound.
The two are highly expressive in conjunction, though neither is sufficient alone.
To do so, we lightly fine-tune a pretrained T2M model with added music event curves, enabling test-time transfer by simply substituting video event curves at inference.
For text conditioning, we extract visual captions from video and summarize them with an LLM to capture the musical style and mood~\cite{vibe}, achieving robust adaptation to diverse and complex scenes.\footnote{
Demos with fixed event-curves but varying text inputs are in the supp. material.
}

\input{figure/teaser_event.tex}

Our contributions are threefold.
First, we propose \ourmodela, the first zero-pair framework for time-synchronized video-to-music generation. Our key insight is that event curves from intra-modal similarity yield structurally comparable representations, enabling test-time transfer from music to video conditioning with only lightweight fine-tuning (192-768 GPU hours) of a pretrained model.
Second, through extensive ablations, we show that \ourmodel is robust across video encoders rather than tied to a specific backbone,
and we analyze design choices that mitigate the modality gap between music and video representations.
Finally, we demonstrate state-of-the-art results across three diverse datasets. Our zero-pair approach surpasses the strongest prior baselines per metric on all metrics across all datasets: \textbf{5--9\%} better in music quality, \textbf{13--15\%} in semantic alignment, \textbf{21--52\%} in temporal synchronization, and \textbf{28\%} in beat alignment on dance videos. A large-scale crowdsourced listening test corroborates these gains.

%% file: figure/teaser_event.tex
\begin{figure}[t!]
    \centering
    \begin{minipage}[t]{0.52\linewidth}
        \centering
        \includegraphics[width=\linewidth]{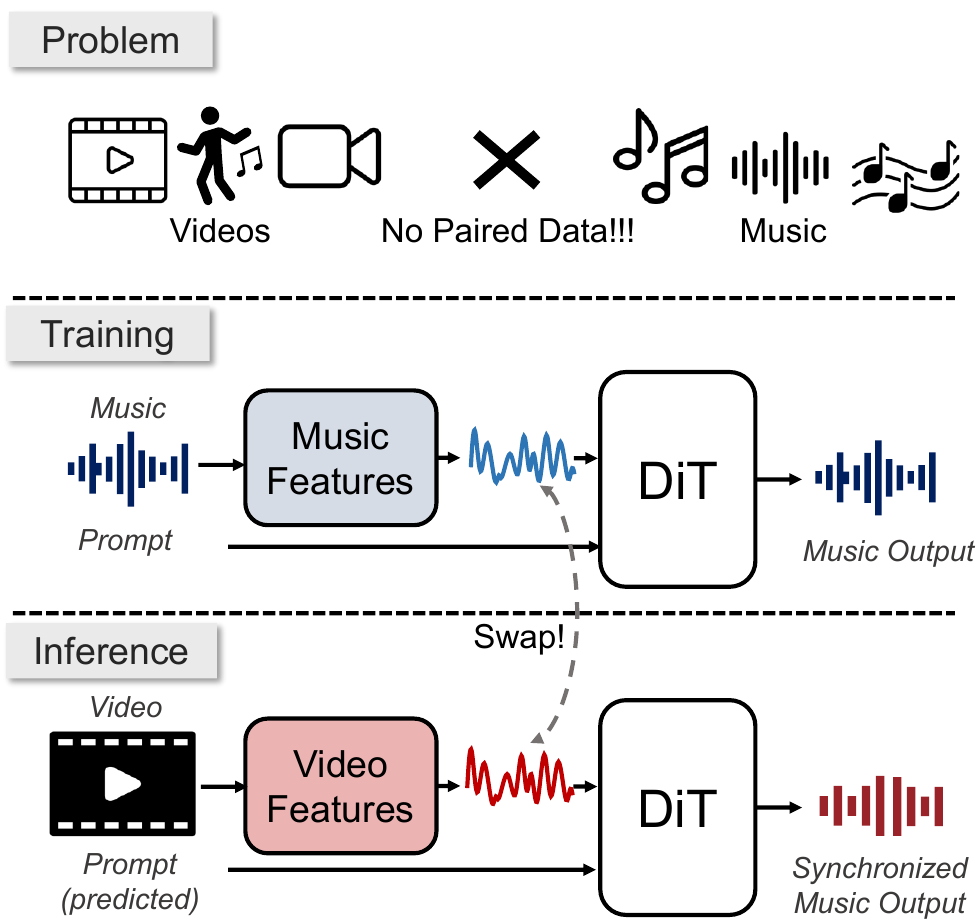}
        \subcaption{
        \textbf{Zero-pair training \& inference.}
        \textbf{Top}: Paired methods require large paired video-music datasets.
        \textbf{Middle}: \ourmodel trains with music-event curve conditioning and does not use paired music and video data.
        \textbf{Bottom}: At inference, we \emph{swap} music-event curves with video-event curves to generate time-synchronized music.}
        \label{fig:teaser}
    \end{minipage}
    \hfill
    \begin{minipage}[t]{0.44\linewidth}
        \centering
        \includegraphics[width=\linewidth]{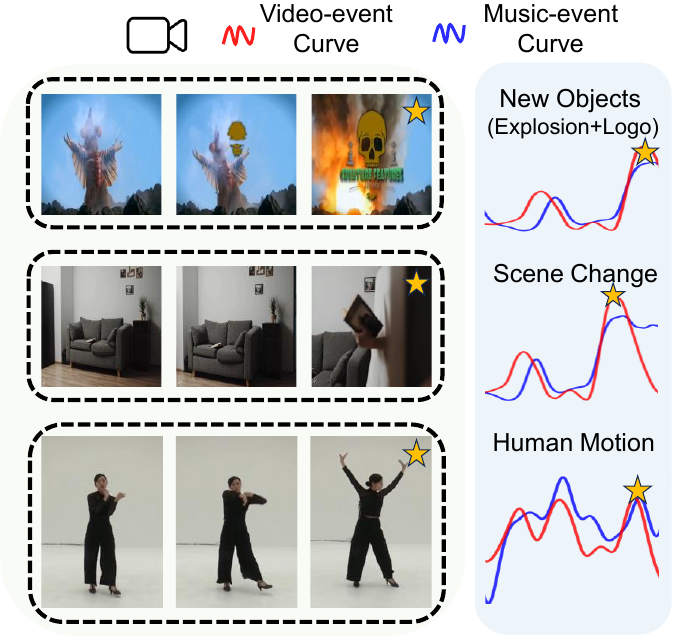}
        \subcaption{
        \textbf{Shared temporal structure across modalities.}
        Real event curves from video and music share similar temporal patterns.
        Ground-truth pairs have correlation $\approx$ \textbf{0.6}; random temporal offsets degrade this to $\approx$ \textbf{0.2}.\protect\footnotemark}
        \label{fig:dynamic}
    \end{minipage}
    \caption{\textbf{Why zero-pair video-to-music generation is possible.} \ourmodel is trained with music-event curves and, at inference, replaces them with video-event curves. This transfer is enabled by the shared temporal structure of event curves across music and video.}
    \vspace{\figmargin}
\end{figure}
\footnotetext{Correlation w/1-sec window around video scene cuts. Correlation degrades as music is time-shifted relative to video.}

%% file: 2_related_work.tex
\section{Related work}
\vspace{\secmargin}

\subsection{Text-to-music generation}
\vspace{\subsecmargin}
General text-to-audio generation has advanced significantly in the past few years~\cite{nips25_thinksound,arxiv23_audiobox,icml23_audioldm,arxiv23_audioldm2,taslp23_audiolm,icml23_make_an_audio,taslp23_diffsound,arxiv24_fastaudio,nips23_codi,arxiv23_codi2,arxiv24_stable_audio,audiogen,arxiv25_fast_t2a,rowles2025foley}.
Text-to-music generation has followed a similar trend~\cite{arxiv25_yue,tian2026audio_omni,arxiv23_musiclm,musicgen,nips23_melody,arxiv23_Mousais,iclr24_MAGNET,tmlr25_t2m,arxiv25_ace_step,icml25_song_gen,arxiv24_flux_music,icml25_ditto}. Both music and audio generation methods typically fall into two modeling paradigms.
The first, autoregressive (AR) modeling~\cite{musicgen,arxiv23_musiclm,audiogen,iclr24_MAGNET,nips23_melody}, operates on discrete audio tokens produced by a neural audio codec~\cite{soundstream,encodec,dac}. A causal transformer predicts each token conditioned on the previously generated tokens and the input text prompt, and the resulting token sequence is decoded back to a waveform by the codec.
The second, latent diffusion models (LDMs)~\cite{arxiv24_flux_music,icml23_audioldm,arxiv23_audioldm2,acmmm23_tango,arxiv23_mustango,icml25_ditto}, learns a denoising process over continuous latents conditioned on text. These employ a similar kind of neural audio codec with continuous latents~\cite{arxiv24_stable_audio}.
While recent text-to-music models~\cite{arxiv23_musiclm,musicgen,nips23_melody,arxiv23_Mousais,iclr24_MAGNET,tmlr25_t2m,arxiv25_ace_step,icml25_song_gen,arxiv24_flux_music} can effectively capture high-level semantics, such as genre, mood, and instrumentation, it remains challenging to align music with fine-grained visual events and/or separate timing and semantic control.

\vspace{\subsecmargin}
\subsection{Video-to-music generation}
\vspace{\subsecmargin}
Building on video-to-audio research~\cite{cvpr23_v2a,cvpr25_multifo,cvpr25_vintage,arxiv24_soundify,arxiv23_foleygen,nips23_diff_foley,cvpr_owens2016visually,eccvw_chen2018visually,cvpr18_visual2sound,tip20_v2a,wang2025kling,liu2024tell,su2024vision,cvpr25_mmaudio}, recent video-to-music methods generate soundtracks aligned with visual rhythm and motions (\ie dance videos)~\cite{nips19_dance2music,zhuang2022music2dance,shlizerman2018audio,eccv22_dnace2music,iclr23_discrete_music,icml23_long_video_music,siggraph24_d2m,cvpr25_enhancing}.
Early approaches relied on symbolic data such as MIDI or ABC notation~\cite{iccv23_v2music,arxiv23_suitable_video2music,acmmm21_cmt,tian2025_xmusic,cvpr24_diffbgm}, but these datasets were limited in scale and expressivity.
Recent works train on paired internet videos accompanied by music tracks~\cite{lin2025vmas,ji2026diff,audiox,cvpr25_vidmuse,arxiv25_extending,ismir25_OES,aaai25_gvmgen,acmmm25_controllable,arxiv24_vidmusician,arxiv24_muvi,cvpr24_melfusion,arxiv23_mu,cvpr25_harmonyset,aaai24_v2meow,tong2025videoechoedmusicsemantic,qi2025customized,gu2026vmsp}.
Such internet data is often noisy, containing imperfect mixing, and/or potential copyright issues.
These limitations motivate the exploration of unpaired or zero-shot paradigms that leverage independent, unpaired music and video sources.

\vspace{\subsecmargin}
\subsection{Video-to-music generation via prompting}
\vspace{\subsecmargin}
Without paired video-music training data, recent methods~\cite{arxiv24_zs_multimodal,aigc25_zs_mozart,arxiv24_zs_m2m,arxiv25_zs_musiscene,icassp25_sonique,cvpr25_filmcomposer} bridge the video and music domains through prompts. Specifically, they translate visual content into text prompts (via LLM reasoning or learned prediction) and then generate music using off-the-shelf text-to-music models.
While this strategy effectively captures high-level semantics (\eg genre, mood), it struggles with fine-grained temporal alignment because such prompts lack the expressiveness to specify timing and dynamics.
In contrast, \ourmodel directly conditions on music or video event curves, enabling temporal synchronization with scene cuts, motion patterns, and dance rhythm.

%% file: 3_Method.tex
\input{figure/fig_method}
\section{Technical approach}
\vspace{\secmargin}
\label{sec:method}
We fine-tune a pretrained T2M latent rectified flow model by augmenting it with temporal event curve conditioning, enabling test-time transfer to video at inference. The curves contribute timing, while semantic content~(energy, genre) comes from text prompts.
Our insight is that temporal change, measured via intra-modal similarity in the feature space of strong encoders, captures perceptually meaningful domain-agnostic event structure.
Events such as beat onsets in music or scene cuts in video both manifest as local dissimilarity between consecutive temporal segments.
By standardizing these dissimilarity signals per-sample, we obtain structurally comparable representations across modalities.
During training, the model is conditioned on text prompts and music-event curves.
At inference, we replace music-event with video-event curves for zero-paired, time-synchronized generation without any model retraining. This alleviates the need for paired video-music data while generating temporally aligned outputs.
An overview is shown in \figref{method}.

\vspace{\subsecmargin}
\subsection{Preliminaries}
\vspace{\subsecmargin}

\noindent\textbf{Music autoencoder.}
We use a pretrained music autoencoder~\cite{casebeer2026generative} to encode stereo music waveforms into continuous latents $\mathbf{x}_0\!\in\!\mathbb{R}^{d\times l}$, where $d$ is the latent dimension and $l$ is the temporal length.
These latents represent the low-level representation used by our rectified flow model.
Separately, we extract high-level semantic features from other pretrained encoders to compute event curves.

\noindent\textbf{Rectified flows.}
Given a text condition $\mathbf{c}$ and a music latent $\mathbf{x}_0$ sampled from the empirical data distribution $\mathcal{P}_D$, a rectified flow model learns a velocity field $f_{\theta}(\cdots)$ that transports samples from a Gaussian prior $\mathcal{N}(\mathbf{0},\mathbf{I})$ to $\mathcal{P}_D$.
The timestep variable $t\!\in\![0,1]$ interpolates between noise ($t=1$) and data ($t=0$) via the linear path $\mathbf{x}_t = t\boldsymbol{\epsilon} + (1 - t)\mathbf{x}_0$, where $\boldsymbol{\epsilon}\!\sim\!\mathcal{N}(\mathbf{0}, \mathbf{I})$.
We train $f_{\theta}(\cdots)$ to predict the velocity $\boldsymbol{\epsilon}-\mathbf{x}_0$ via:
\vspace{\eqmargin}
\begin{equation}
\label{eq:rf_loss}
\min_{\theta}\;\mathbb{E}_{\mathbf{x}_0,\boldsymbol{\epsilon},t,\mathbf{c}}
\left\|\,(\boldsymbol{\epsilon}-\mathbf{x}_0) - f_{\theta}(\mathbf{x}_t,\mathbf{c},t)\,\right\|_2^2.
\end{equation}
We generate samples by solving the ODE
$d\mathbf{x}_t = -f_{\theta}(\mathbf{x}_t,\mathbf{c},t)\,dt$ from $t=1$ to $t=0$ with 96 sampling steps and classifier-free guidance~\cite{cfg}.

\noindent\textbf{Model architecture.}
We use a pretrained Diffusion Transformer (DiT) architecture~\cite{peebles2023scalable} $f_{\theta}(\cdots)$ with cross-attention text conditioning $\mathbf{c}$ following standard T2M models~\cite{arxiv24_flux_music,icml23_audioldm,arxiv23_audioldm2,acmmm23_tango,arxiv23_mustango,arxiv24_stable_audio}.

\vspace{\subsecmargin}
\subsection{Temporal event curves}
\label{subsec:ssm}
\vspace{\subsecmargin}
We construct event curves that capture when and how much meaningful change occurs (sometimes denoted as novelty curves).
Our procedure is identical for music and video, enabling direct substitution at inference.

\noindent\textbf{Perspectives on event curves.}
Our event curve formulation connects to established uses of self-similarity across modalities.
In music, self-similarity has been widely used to analyze musical structure and rhythm~\cite{foote2001visualizing, peeters2023self} and is also considered a generalization of the concept of autocorrelation.
In video, similar approaches segment footage and detect natural boundaries such as shot transitions~\cite{shechtman2007matching, kwon2021learning}
Mathematically, our approach is equivalent to extracting one or more off-diagonal bands of the self-similarity matrix.
In that sense, our formulation leverages the natural structure of self-similarity to enable zero-shot transfer.

\noindent\textbf{Features for event curves.}
To compute event curves, we begin by extracting temporal feature sequences using pretrained encoders.
For music (training only), we apply a music encoder to obtain $\mathbf{f}_m  \in \mathbb{R}^{d_m \times l_m}$, where $l_m$ is the temporal length and $d_m$ is the feature dimension.
For video (inference only), we encode each frame with a visual encoder and spatially pool to obtain $\mathbf{f}_v \in \mathbb{R}^{d_v \times l_v}$, where $l_v$ is the number of frames and $d_v$ is the feature dimension.
These high-level semantic features are used solely to compute event curves.

\noindent\textbf{Computing event curves.}
Given a feature sequence $\mathbf{f} \in \mathbb{R}^{d_f \times l_f}$ (either $\mathbf{f}_m$ or $\mathbf{f}_v$), where $\mathbf{f}^k$ denotes the $k-th$ temporal feature-vector, we measure temporal change via the cosine similarity between consecutive vectors:
\vspace{\eqmargin}
\begin{equation}
\label{eq:cos_sim}
s^k =
\frac{\mathbf{f}^k \cdot \mathbf{f}^{k+1}}
{\|\mathbf{f}^k\|\,\|\mathbf{f}^{k+1}\|},
\quad k = 1, \dots, l_f{-}1.
\end{equation}
We obtain a dissimilarity sequence via $a^k = 1 - s^k$ and set $\mathbf{A}=\{a^k\}$. Higher values indicate stronger temporal change in high-level encoder spaces, including beat onsets, scene cuts, motion, and musical-structure changes. We provide preliminary analysis of more elaborate self-similarity schemes in Appendix~\ref{sec:additional_results}, Tab.~\ref{tab:sd_audio_appendix}.

\noindent\textbf{Mitigating the modality gap.}
To enable zero-shot transfer from music to video, we apply three operations that align event curves across modalities.
First, we standardize each datum $\mathbf{A}$ to have zero mean and unit variance:
\vspace{\eqmargin}
\begin{equation}
\label{eq:standard_norm}
\bar{a}^k = \frac{a^k - \mu(\mathbf{A})}{\sigma(\mathbf{A})}.
\end{equation}
Without standardization, music and video event curves have different scales and offsets, creating a distribution shift when swapping at inference.
This intentionally makes the curve a relative temporal-change signal. Absolute energy, mood, and instrumentation are instead controlled by the text prompt.
Second, we resample to length $l$ (matching the temporal dimension of $\mathbf{x}_0$).
Third, we apply temporal smoothing (e.g. via a Hann window) to suppress modality-specific details while preserving the larger structure, yielding the final event curve:
\begin{equation}
\label{eq:event_curve}
\mathbf{e} = \text{Smooth}\!\left(\text{Resample}\!\left(\bar{\mathbf{A}}, l\right)\right)
\;\in\;\mathbb{R}^{l}.
\end{equation}
During training, we compute $\mathbf{e}_m$ from music features. At inference, we compute $\mathbf{e}_v$ from video features.
Our design is agnostic to the choice of feature encoders. We use MusicFM~\cite{musicfm} and DINOv2~\cite{dinov2} by default but explore alternative audio and video encoders in Section~\ref{sec:abs}.

\vspace{\subsecmargin}
\subsection{Rectified flow model fine-tuning}
\vspace{\subsecmargin}
\label{subsec:cond}
We incorporate event curves into a pretrained text-conditioned rectified flow model via fine-tuning.

\noindent\textbf{Event curve conditioning via concatenation.}
We inject the event curve $\mathbf{e}$ by concatenating it as an additional channel to the rectified flow latent:
\vspace{\eqmargin}
\begin{equation}
\label{eq:cond_concat_no_repeat}
\widetilde{\mathbf{x}}_t
=
\big[\;\mathbf{x}_t\;,\;\mathbf{e}\;\big]
\;\in\;\mathbb{R}^{(d{+}1)\times l},
\end{equation}
where $[\cdot,\cdot]$ denotes channel-wise concatenation.
This approach is simple and parameter-efficient, only requiring additional parameters in the input projection layer of the DiT.
The same concatenation approach also supports event curves computed from different temporal scales, as explored in Appendix~\ref{sec:additional_results}.

\noindent\textbf{Fine-tuning objective.}
We initialize from a pretrained T2M model, add the conditioning signal, and fine-tune it with music-event curve conditioning via:
\vspace{\eqmargin}
\begin{equation}
\label{eq:cond_loss_final}
\min_{\theta}\;\mathbb{E}_{\mathbf{x}_0,\boldsymbol{\epsilon},t,\mathbf{e}_m,\mathbf{c}}
\left\|\,(\boldsymbol{\epsilon}-\mathbf{x}_0)-
f_{\theta}\!\big(\widetilde{\mathbf{x}}_t,\mathbf{c},t\big)\,\right\|_2^2,
\end{equation}
where we explicitly denote $\mathbf{e}_m$ to emphasize that during training, event curves are computed from music.
Given a music clip during training, we extract music encoder features $\mathbf{f}_m$ and compute the music event curve $\mathbf{e}_m$ via the procedure in Section \ref{subsec:ssm}.
The text condition $\mathbf{c}$ is obtained from ground-truth music descriptions in the pre-training data.
Fine-tuning allows our model to learn to generate music conditioned on both text and temporal events.

\subsection{Zero-pair inference via event curve swapping}
\vspace{\subsecmargin}
\label{subsec:vs}

At inference, we perform test-time transfer to video-to-music generation by swapping in video event curve conditioning signals, while keeping all model weights fixed.
Given an input video, we extract visual encoder features $\mathbf{f}_v$ and compute the video event curve $\mathbf{e}_v$ using the same procedure as $\mathbf{e}_m$ .
For text conditioning, we use an LLM~\cite{vibe} to generate and summarize video captions into a music-appropriate text prompt $\mathbf{c}$.
We then generate music using standard flow-model inference and add the generated audio to the input video.
Note, our music and video event curves are designed to mitigate the domain gap, so this swapping does not require retraining.

%% file: figure/fig_method.tex
\begin{figure*}[t!]
    \centering
    \includegraphics[width=0.9\linewidth]{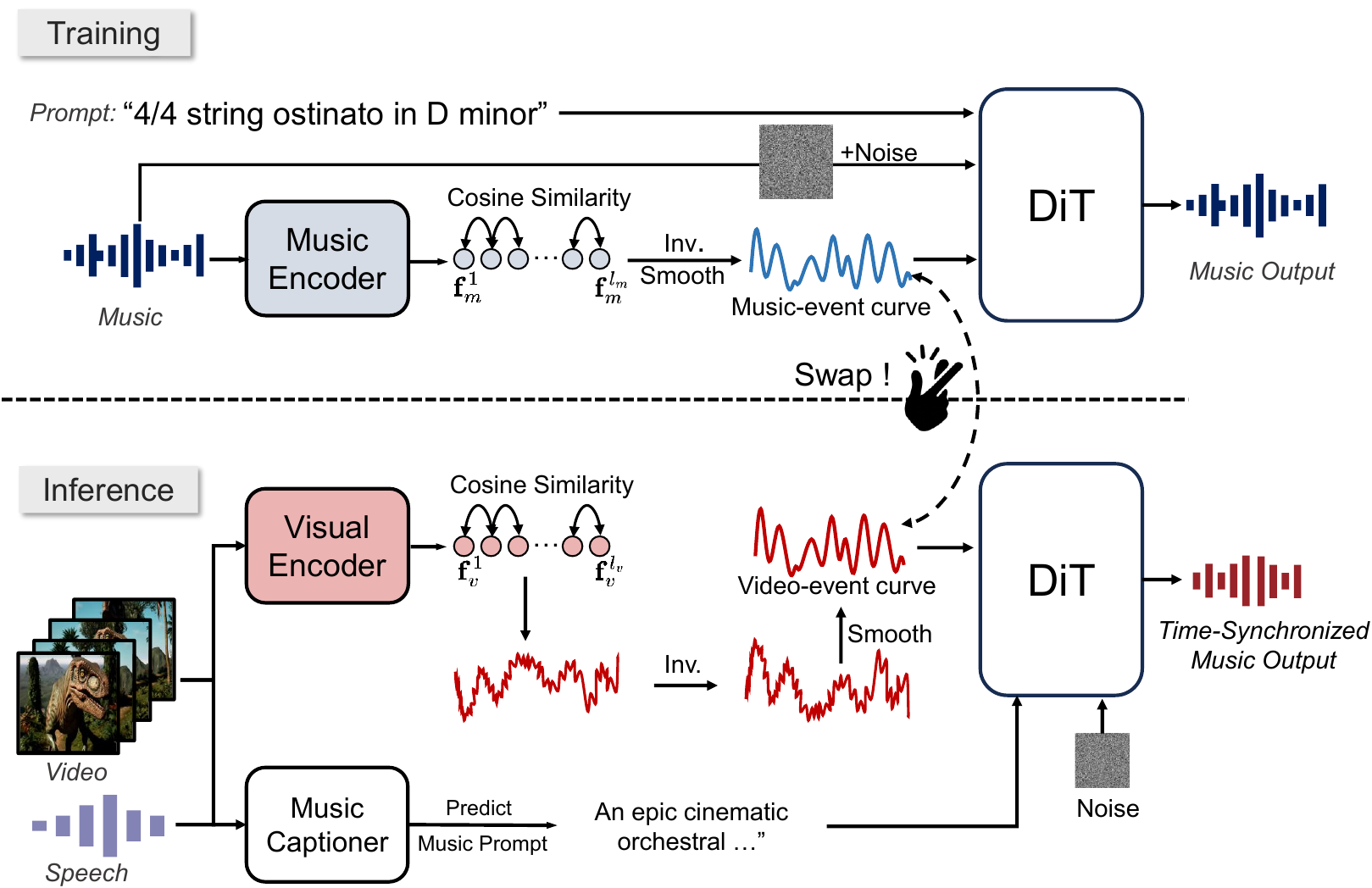}
    \caption{\textbf{Method overview.}
    \textbf{Top}: During training, \ourmodel\ learns a rectified-flow diffusion process conditioned on text prompts and a \emph{music-event curve} derived from intra-music similarity.
    \textbf{Bottom}: At inference, music conditioning is \emph{swapped} with a \emph{video-event curve} based on framewise similarity, enabling zero-pair, time-synchronized video-to-music generation.
    For semantic alignment, a text prompt is predicted from video and speech via Vibe~\cite{vibe}, which first summarizes video content and recommends a matching music text prompt.
    }
    \vspace{\figmargin}
    \label{fig:method}
\end{figure*}

%% file: 4_result.tex
\section{Experimental setup}
\vspace{\secmargin}
\label{sec:exp_setup}
We use three diverse benchmark datasets, multiple baselines, and a standard protocol for evaluation.

\vspace{\subsecmargin}
\subsection{Evaluation data}
\vspace{\subsecmargin}
We test on datasets spanning general, cinematic, and dance videos:
{\setlength{\leftmargini}{2.0em}
\begin{compactitem}
  \item \textbf{OES-Pub}~\cite{ismir25_OES} is the public evaluation split of the Open Screen Soundtrack Library (OSSL), containing $115$ public-domain movie clips paired with royalty-free music. Each clip is $\approx 30$ seconds and includes human-annotated music prompts.
  \item \textbf{MovieGenBench-Music}~\cite{moviegen} is the music subset of MovieGenBench, consisting of $527$ video-music pairs with sound effects across diverse video types. Each clip is $\approx 10$ seconds and includes a music prompt.
  \item \textbf{AIST++}~\cite{aist++,aist} is a street dance dataset with copyright-cleared dance music, consisting of $20$ video–music pairs across $10$ dance genres. Each clip is $\approx 7$ seconds and includes BPM.
\end{compactitem}
Collectively, these benchmarks cover the principal video categories evaluated in prior video-to-music work: natural, cinematic, and dance.

\vspace{\subsecmargin}
\subsection{Evaluation metrics}
\label{sec:evaluation}
\vspace{\subsecmargin}
Following \cite{cvpr25_vidmuse,audiox,lin2025vmas,aaai25_gvmgen}, we measure: audio fidelity, semantic alignment, and temporal synchronization. For each dataset under test, we customize the metrics used to highlight the critical aspects of the specific video-to-music generation task.

\noindent\textbf{OES-Pub \& MovieGenBench-Music:}
\begin{compactitem}
    \item \textbf{Fréchet audio distance (FAD)}~\cite{fad}: audio fidelity via distributional distance between reference and generated music in VGGish~\cite{vggish} space. We denote the use of a separate reference set~\cite{song} via *. Lower is better.
    \item \textbf{CLAP score}~\cite{clap}: semantic alignment via cosine similarity between generated music and text prompts in CLAP space. Higher is better.
    \item \textbf{Scene cut hit (SCH)}\footnote{Pseudocode in appendix.}: aperiodic temporal alignment. A \emph{hit} occurs when a beat falls within $\pm100$\,ms of a scene cut, computed as $\text{hits}/\text{total cuts}$. Inspired by \cite{lin2025vmas}. Higher is better.
    \item \textbf{Human evaluation}: measures subjective preference for music quality and video synchronization.
    Given two music tracks for the same video, raters answer: 1) \textit{Which has better quality?} and 2) \textit{Which better synchronizes with visuals?}
    We collected 1403 valid votes from the crowd-source platform Appen, evenly sampled across datasets and randomly sampled across baselines
    We report win-rates with pairwise t-tests vs.~\cite{arxiv24_mumu,aaai25_gvmgen,cvpr25_vidmuse,audiox,icassp25_sonique,acmmm25_controllable}. Higher is better.
\end{compactitem}

\noindent\textbf{AIST++:}
\begin{compactitem}
    \item \textbf{Beat coverage (BCS), beat hit score (BHS), F1}: periodic rhythm alignment~\cite{siggraph24_d2m}. BCS and BHS are recall and precision of music beats relative to motion beats, and F1 is their harmonic mean. Higher is better.
    \item \textbf{Temporal deviation (TD)}: tempo difference from ground truth. We use $.2$s tolerance (reduced from $1.0$s) for perceptual meaningfullness~\cite{siggraph24_d2m}. Lower is better.
\end{compactitem}
}

\input{table/sota}
\section{Results and analysis}
\vspace{\secmargin}
\subsection{Comparison with state of the art}
\vspace{\subsecmargin}
\ourmodel demonstrates strong performance across all benchmarks, outperforming both paired/unpaired baselines despite using no paired video-music training data. We analyze results across audio fidelity, semantic alignment, and temporal synchronization, highlighting key insights on cross-domain generalization and the utility of event-curves.

\textbf{General \& cinematic video to music generation.}
\tabref{sota} shows results on OES-Pub and MovieGenBench-Music with DINOv2~\cite{dinov2} as the visual encoder.
\ourmodel achieves the best audio quality on both benchmarks (\ie FAD*: 4.95 on OES-Pub; FAD: 2.68 on MovieGenBench).
Notably, our method exhibits the most consistent performance across datasets, while paired baselines show substantial variation (\eg VidMuse: 10.4 $\rightarrow$ 2.98).
Interestingly, VidMuse, trained on 18k hours of paired data, achieves the worst FAD on OES-Pub but the best among paired methods on MovieGenBench, suggesting dataset-specific overfitting rather than robust generalization.
We compute FAD* on OES-Pub using SongDescriber~\cite{song} as the reference distribution, since the original OES-Pub audio contains speech, sound effects, and background noise.
MovieGenBench uses generated music references, which generally yield lower absolute FAD scores than real recordings.
For semantic alignment (CLAP), paired-data baselines perform reasonably on OES-Pub but exhibit noticeable degradation on MovieGenBench-Music.
For instance, AudioX achieves 0.08 on MovieGenBench-Music, compared to 0.19 on OES-Pub.
This suggests limited generalization to MovieGenBench's generated videos, which differ from the real paired data used during training.
Interestingly, SONIQUE is the only baseline that improves on MovieGenBench (\ie 0.09 to 0.16), likely because its pure-text approach is less sensitive to domain shift in the visual content.
In contrast, \ourmodel demonstrates robust performance across all datasets (0.23 on OES-Pub, 0.18 on MovieGenBench-Music), showing cross-domain generalization with better CLAP scores.

For temporal alignment (SCH), methods trained on paired video-music data achieve strong performance, ranging from 0.33--0.40 on OES-Pub and 0.24--0.48 on MovieGenBench-Music.
In comparison, SONIQUE, which uses a pure LLM-based prompting approach, achieves lower temporal alignment (\ie 0.27 on OES-Pub, 0.21 on MovieGenBench-Music).
This demonstrates the benefit of learning from paired data for capturing temporal structure.
However, \ourmodel achieves strong temporal alignment (\ie 0.61 on OES-Pub, 0.58 on MovieGenBench-Music), showing that explicit event-curve conditioning can surpass paired supervision for fine-grained synchronization.

\input{table/sota_human_eval}

\textbf{Human evaluation.}
We analyze our 1403 pairwise human ratings across six baselines and two questions via multiple pairwise $t$-tests with Bonferroni correction~\cite{holm1979simple}. All win-rates reported in \tabref{human_eval} show our win-rate (higher is better). We show results across all videos in \tabref{human_eval_a} and videos with a scene cut in \tabref{human_eval_b}.
Our method is consistentley preferred when compared head-to-head in win-rate against each alternative.
On general videos, our approach has a statistically significant win-rate over all baselines in music quality. We see a similar trend in temporal alignment, where our method is best in 6/6 cases and the lead is significant in 5/6.
To understand when and why \ourmodel is preferred, we test scenes with a scene-cut in \tabref{human_eval_b}, a clear event for raters to anchor their evaluation on. Average win-rate increases and is significant in all 12/12 cases.
V2M-Zero is preferred over paired-data baselines and this preference increases with an anchoring visual event.

\subsection{Generalization across video types and models}
\vspace{\subsecmargin}

\textbf{Dance video to music generation.}
To evaluate on content with tightly coupled temporal dynamics, we test on AIST++~\cite{aist,aist++}, a dance dataset where video and music are highly related.
\tabref{dance} shows results comparing against methods explicitly designed or trained for dance-to-music generation.
\ourmodel achieves strong performance across all rhythm metrics (0.5818/0.6274/0.5856/12.24), outperforming specialized methods.
Improvements are most pronounced for BHS and TD, suggesting that event curve conditioning achieves precise and accurate motion-rhythm correspondence.
By decoupling event-curve extraction from music generation, the same model can be used on diverse video domains without retraining.

\begin{wrapfigure}{r}{0.40\linewidth}
    \vspace{-15mm}
    \centering
    \includegraphics[width=0.99\linewidth]{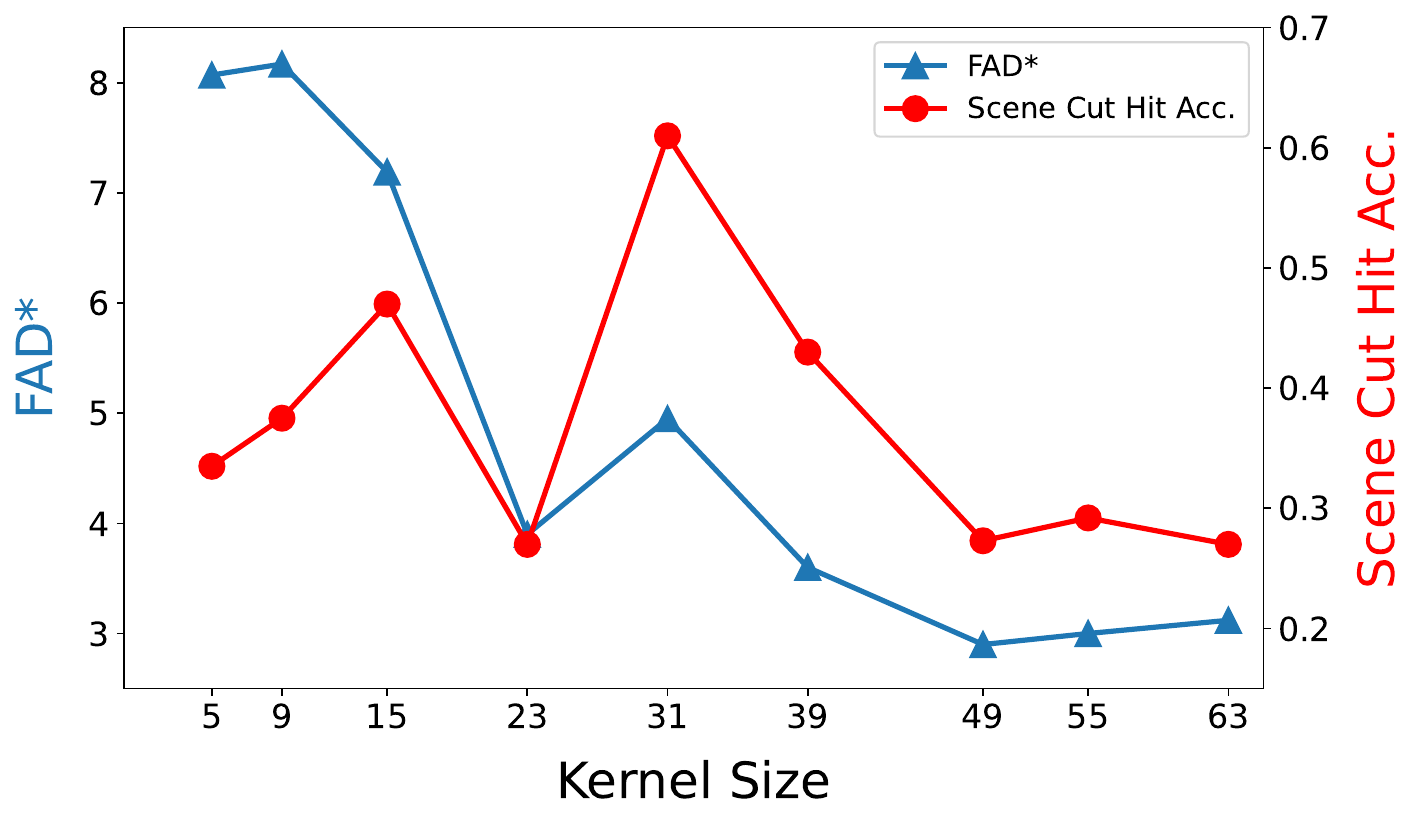}
    \vspace{-5mm}
    \caption{
    \textbf{Smoothing kernel size sweep on OES-Pub with MusicFM and DINOv2:} audio quality (FAD*) improves with larger kernels; synchronization (SCH) peaks at intermediate sizes. We hyperparameter tune the kernel size and apply a standard (31) across all other datasets and encoders.
    }
    \vspace{-15mm}
    \label{fig:gap}
\end{wrapfigure}

\input{table/sota_dance}
\noindent\textbf{Generalization to public text-to-music models.}
To confirm the generality of \ourmodel, we evaluate with a pre-trained open text-to-music model, Stable-Audio-ControlNet~\cite{sd_audio_ctrl}.
This model is trained with audio RMS curves, which we swap for video-event curves at inference time.
\tabref{sd_audio} has comparable FAD* and CLAP, but improved SCH (0.28 to 0.38).
The improvement in temporal alignment, via zero-shot integration, shows that event-curve conditioning is model-agnostic.

\input{table/encoder_llm}

\vspace{\subsecmargin}
\subsection{Ablation studies}
\label{sec:abs}
\vspace{\subsecmargin}
We systematically ablate four design axes: (i) kernel size for modality gap mitigation, (ii) encoders for event-curve extraction, and (iii) LLM selection for prompt generation.

\textbf{Mitigating modality gap.}
Music-event curves (training) and video-event curves (inference) differ in temporal granularity, creating a modality gap.
We mitigate this via Hann-window smoothing applied to both modalities. We sweep kernel size on OES-Pub with the MusicFM and DINOv2 encoders, then fix this value for all other datasets and encoders without re-tuning.
In \figref{gap} we find that increasing the kernel from 9 (.7 seconds) to 63 (5 seconds) improves FAD: 8.17 to 3.12.
However, excessive smoothing blurs fine-grained events, causing SCH to degrade: 0.61 to 0.27.
This reveals a trade-off. Stronger smoothing improves audio quality but weakens temporal alignment.
This matches intuition as music is not frame-by-frame synchronized to video unless for comedic effect, known as Mickey Mousing~\cite{wegele2014max}.
We pick kernel size 31 to balance the two.

\textbf{Encoder architecture for event curves.}
\tabref{encoder} compares three encoder design paradigms: shared, reconstruction, and self-supervised encoders.

\noindent\textbf{Shared encoders.}
We test whether shared-weight audio-visual encoders, like AVSiam~\cite{avsiam}, can reduce the modality gap by embedding both modalities in a common space.
AVSiam achieves the best FAD (4.52) among all configurations, suggesting improved distribution matching.
However, temporal alignment degrades significantly (SCH: 0.61 $\rightarrow$ 0.35), likely due to AVSiam's lower specialized capacities in both the audio and visual domains compared to foundation models like DINOv2~\cite{dinov2} and a fundamental difference on how music and audio are aesthetically matched to music.
We hypothesize that large-scale shared encoders could close this gap as a promising future direction.

\noindent\textbf{Music encoder.}
The music encoder has the strongest impact on performance.
We initially use the same encoder as the latent diffusion VAE, for simplicity.
Using a self-supervised model like MusicFM~\cite{musicfm} improves temporal synchronization (SCH: 0.31 $\rightarrow$ 0.61, with DINOv2), while also improving audio quality (FAD: 4.95 $\rightarrow$ 4.77).
MusicFM's more semantic representations better correlate with visual event patterns, enabling more reliable curve alignment during inference.

\noindent\textbf{Video encoder.}
We compare self-supervised visual encoders: V-JEPA~\cite{vjepa2} and DINOv2~\cite{dinov2}.
DINOv2 yields slightly better audio quality across music encoders (FAD: 4.77 vs. 5.13 with VAE; 4.95 vs. 5.02) with MusicFM~\cite{musicfm}).
However, alignment results are mixed: V-JEPA outperforms with VAE (SCH: 0.41 vs. 0.31), while DINOv2 pairs best with MusicFM (SCH: 0.61 vs. 0.48).
Overall, video encoder choice has minimal impact; alignment quality depends more critically on the music encoder.
We use MusicFM + DINOv2 as our default configuration.

\textbf{LLM selection for music prompt generation.}
\tabref{llm} compares three LLMs for generating music prompts from video: Gemma-4B~\cite{team2025gemma}, Qwen3-4B~\cite{qwen3}, and Llama-3.2-3B~\cite{llama3}.
Event curves are fixed, and we ablate the prompts.
Differences across LLMs are minimal: FAD, CLAP, and SCH vary by less than $5\%$.
Gemma-4B is marginally better (FAD: 4.95, CLAP: 0.23, SCH: 0.61).
We conclude that LLM choice has a negligible impact and any modern LLM suffices for semantic guidance.

%% file: table/sota.tex
\begin{table*}[!t]
\setlength{\tabcolsep}{4.5pt}
\caption{\textbf{Comparison with state of the art.} We evaluate all methods~\cite{arxiv24_mumu,aaai25_gvmgen,cvpr25_vidmuse,audiox,acmmm25_controllable,icassp25_sonique} on OES-Pub~\cite{ismir25_OES} and MovieGenBench-Music~\cite{moviegen} in audio quality (FAD), high-level semantic alignment (CLAP), and our proposed temporal metric (SCH). \ourmodel outperforms all competing methods across all datasets and metrics. Note, OES-Pub reference data includes non-musical sounds, so we use SongDescriber~\cite{song} for the reference distribution for FAD results.}
\centering
\resizebox{0.98\linewidth}{!}{
\begin{tabular}{l|cc|ccc|ccc}
\toprule
\multicolumn{1}{c}{}& & &
\multicolumn{3}{c|}{\textbf{OES-Pub.}} &
\multicolumn{3}{c}{\textbf{MovieGenBench-Music}} \\
\cmidrule(lr){4-6}\cmidrule(lr){7-9}

\multicolumn{3}{c|}{} &
\multicolumn{1}{c}{\makecell{Audio \\ Quality}} &
\multicolumn{1}{c}{\makecell{High-Level \\ Alignment}} &
\multicolumn{1}{c|}{\makecell{Low-Level \\ Matching}} &
\multicolumn{1}{c}{\makecell{Audio \\ Quality}} &
\multicolumn{1}{c}{\makecell{High-Level \\ Alignment}} &
\multicolumn{1}{c}{\makecell{Low-Level \\ Matching}} \\
\cmidrule(lr){4-4}
\cmidrule(lr){5-5}
\cmidrule(lr){6-6}
\cmidrule(lr){7-7}
\cmidrule(lr){8-8}
\cmidrule(lr){9-9}

\multicolumn{1}{l|}{\textbf{Method}} &
\multicolumn{2}{c|}{\textbf{V–M Pairs}} &
\textbf{FAD*}$\downarrow$ &
\textbf{CLAP}$\uparrow$ &
\textbf{SCH}$\uparrow$ &
\textbf{FAD}$\downarrow$ &
\textbf{CLAP}$\uparrow$ &
\textbf{SCH}$\uparrow$ \\
\midrule

M$^{2}$UGen~\cite{arxiv24_mumu}         & \bluecheck & 36.7h  & 6.67 & 0.07 & 0.35 & 5.84 & 0.02 & 0.24 \\
GVMGen~\cite{aaai25_gvmgen}             & \bluecheck & 147h   & 6.25 & 0.18 & 0.35 & 3.96 & 0.06 & 0.48 \\
MTCV2M~\cite{acmmm25_controllable}      & \bluecheck & 147h   & 5.44 & 0.20 & 0.37 & 4.02 & 0.16 & 0.29 \\
VidMuse~\cite{cvpr25_vidmuse}           & \bluecheck & 18000h & 10.4 & 0.16 & 0.40 & 2.98 & 0.04 & 0.47 \\
AudioX~\cite{audiox}                    & \bluecheck & 15793h & 7.46 & 0.19 & 0.33 & 2.82 & 0.08 & 0.30 \\

\midrule
SONIQUE~\cite{icassp25_sonique} & \bluexmark & 0h & 6.80 & 0.09 & 0.27 & 6.47 & 0.16 & 0.21 \\

\midrule
\ourmodel (Ours) & \bluexmark & 0h &
\textbf{4.95} & \textbf{0.23} & \textbf{0.61} &
\textbf{2.68} & \textbf{0.18} & \textbf{0.58} \\

\bottomrule
\end{tabular}
}
\label{tab:sota}
\vspace{-4mm}
\end{table*}

%% file: table/sota_human_eval.tex
\begin{table}[t]
    \centering
    \caption{\textbf{Human evaluation.}
    Pairwise win rates of our model against each baseline (1403 ratings, Bonferroni-corrected $t$-tests, $p < 0.0083$).
    $^\dagger$~above chance but not significant after correction.
    Scene cut subset comprises 67\% of all votes.
    }
    \label{tab:human_eval}

    \begin{minipage}[t]{0.48\linewidth}
        \vspace{0pt}
        \centering
        \subcaption{All Videos}\label{tab:human_eval_a}
        \resizebox{0.99\linewidth}{!}{
        \begin{tabular}{l|cc}
        \toprule
        & \multicolumn{2}{c}{\bf Win Rate of Ours $\uparrow$} \\
        \cmidrule(lr){2-3}
        \bf Baseline & \bf Music Quality & \bf Temp.\ Align. \\
        \midrule
        M$^2$UGen & 67.45\%               & 66.51\%            \\
        GVMGen    & 67.00\%               & 67.00\%            \\
        MTCV2M    & 59.72\%               & 59.72\%  \\
        VidMuse   & 68.85\%               & 53.77\%$^\dagger$  \\
        AudioX    & 72.36\%               & 60.36\%  \\
        SONIQUE   & 77.16\%               & 73.60\%        \\
        \midrule
        \bf Average & \bf 68.76\%         & \bf 63.49\%        \\
        \bottomrule
        \end{tabular}
        }
    \end{minipage}
    \hfill
    \begin{minipage}[t]{0.48\linewidth}
        \vspace{0pt}
        \centering
        \subcaption{Scene Cut Videos}\label{tab:human_eval_b}
        \resizebox{0.99\linewidth}{!}{
        \begin{tabular}{l|cc}
        \toprule
        & \multicolumn{2}{c}{\bf Win Rate of Ours $\uparrow$} \\
        \cmidrule(lr){2-3}
        \bf Baseline & \bf Music Quality & \bf Temp.\ Align. \\
        \midrule
        M$^2$UGen & 67.88\%                & 66.42\%            \\
        GVMGen    & 70.29\%                & 67.39\%            \\
        MTCV2M    & 65.35\%                & 66.14\%            \\
        VidMuse   & 74.63\%                & 59.51\%  \\
        AudioX    & 73.17\%                & 61.46\%  \\
        SONIQUE   & 81.54\%                & 78.46\%        \\
        \midrule
        \bf Average & \bf 71.14\%         & \bf 66.56\%        \\
        \bottomrule
        \end{tabular}
        }
    \end{minipage}
    \vspace{\tabmargin}
    \vspace{-2mm}
\end{table}

%% file: table/sota_dance.tex
\begin{table}[t]
    \centering
    \caption{\textbf{Generalization across video types and model implementations.}
    (\textbf{a}) Eval. on ~\cite{aist,aist++} using beat consistency (BCS, BHS, F1) and temporal deviation (TD).
    (\textbf{b}) \ourmodela~on a public text-to-music model evaluated on OES-Pub.
    }

    \begin{minipage}[t]{0.55\linewidth}
        \vspace{0pt}
        \centering
        \subcaption{Dance Video Results on AIST++.}
        \label{tab:dance}
        \resizebox{0.99\linewidth}{!}{
        \begin{tabular}{l|cccc}
        \toprule
        \bf Method & \bf BCS$\uparrow$ & \bf BHS$\uparrow$ & \bf F1$\uparrow$ & \bf TD$\downarrow$ \\
        \midrule
        CMT~\cite{acmmm21_cmt}              & 0.3368 & 0.1515 & 0.2090 & 21.74 \\
        CDCD~\cite{iclr23_discrete_music}   & 0.4233 & 0.2151 & 0.2852 & 19.25 \\
        LORIS~\cite{icml23_long_video_music}& 0.3721 & 0.3371 & 0.3537 & 17.80 \\
        MDM~\cite{siggraph23_motion2dance}  & 0.3798 & 0.4185 & 0.3982 & 22.96 \\
        Text Inv.~\cite{siggraph24_d2m}     & 0.4761 & 0.4398 & 0.4572 & 20.34 \\
        \midrule
        \ourmodel (Ours) & \bf0.5818 & \bf0.6274 & \bf0.5856 & \bf12.24 \\
        \bottomrule
        \end{tabular}
        }
    \end{minipage}
    \hfill
    \begin{minipage}[t]{0.44\linewidth}
        \vspace{0pt}
        \centering
        \subcaption{Cross-architecture generalization (OES-Pub).}
        \label{tab:sd_audio}
        \resizebox{0.99\linewidth}{!}{
        \begin{tabular}{lccc}
        \toprule
        \bf Method & \bf FAD*$\downarrow$ & \bf CLAP$\uparrow$ & \bf SCH$\uparrow$ \\
        \midrule
        SD-Audio-ctrl~\cite{sd_audio_ctrl} & 4.86 & \textbf{0.18} & 0.28 \\
        +\ourmodela (Ours) & \textbf{4.13} & 0.17 & \textbf{0.38} \\
        \bottomrule
        \end{tabular}
        }
    \end{minipage}
    \vspace{\tabmargin}
    \vspace{-2mm}
\end{table}

%% file: table/encoder_llm.tex
\begin{table}[t]
\centering
\caption{
  \textbf{(a) Encoder selection for event curves.}
  Impact of music (training) and visual (inference) encoders on OES-Pub.
  MusicFM\,+\,DINOv2 achieves the best balance.
  \textbf{(b) LLM selection for music prompts.}
  Impact of LLM choice for music prompt generation on OES-Pub.
  \ourmodel uses Vibe~\cite{vibe} based on Gemma-4B~\cite{team2025gemma}.
}
\vspace{4pt}
\begin{minipage}[t]{0.54\linewidth}
  \centering
  \subcaption{}
  \label{tab:encoder}
  \resizebox{\linewidth}{!}{
  \begin{tabular}{l c c c c}
    \toprule
    \textbf{\makecell{Music Enc.\\(Training)}} &
    \textbf{\makecell{Visual Enc.\\(Inference)}} &
    \textbf{FAD*}$\downarrow$ &
    \textbf{CLAP}$\uparrow$ &
    \textbf{SCH}$\uparrow$ \\
    \midrule
    \multicolumn{2}{c}{AVSiam~\cite{avsiam}} &
    \textbf{4.52} & 0.19 & 0.35 \\
    \midrule
    VAE~\cite{casebeer2026generative} & V-JEPA 2~\cite{vjepa2} &
    5.13 & 0.18 & 0.41 \\
    VAE~\cite{casebeer2026generative} & DINOv2~\cite{dinov2} &
    4.77 & 0.16 & 0.31 \\
    \midrule
    MusicFM~\cite{musicfm} & V-JEPA 2~\cite{vjepa2} &
    5.02 & 0.18 & 0.48 \\
    MusicFM~\cite{musicfm} & DINOv2~\cite{dinov2} &
    4.95 & \textbf{0.23} & \textbf{0.61} \\
    \bottomrule
  \end{tabular}
  }
\end{minipage}
\hfill
\begin{minipage}[t]{0.45\linewidth}
  \centering
  \subcaption{}
  \label{tab:llm}
  \resizebox{\linewidth}{!}{
  \begin{tabular}{l c c c}
    \toprule
    \textbf{Music Captioner} &
    \textbf{FAD*}$\downarrow$ &
    \textbf{CLAP}$\uparrow$ &
    \textbf{SCH}$\uparrow$ \\
    \midrule
    Qwen3-4B~\cite{qwen3}      & 4.98 & \textbf{0.23} & 0.58 \\
    Llama-3.2-3B~\cite{llama3} & 5.02 & 0.21          & 0.60 \\
    \midrule
    Gemma-4B~\cite{team2025gemma} & \textbf{4.95} & \textbf{0.23} & \textbf{0.61} \\
    \bottomrule
  \end{tabular}
  }
\end{minipage}
\vspace{\tabmargin}
\end{table}

%% file: 5_Conclusion.tex
\section{Conclusions}
\vspace{\secmargin}
\label{sec:conclusions}
We introduce \ourmodela, a video-to-music generation approach that produces time-aligned music with disentangled time synchronization and semantic control from video while requiring zero video-music pairs at training time.
Our observation is that while musical and visual events differ semantically, they exhibit similar temporal patterns when embedded by pretrained music and video encoders.
We exploit this property via event curves, signals computed from consecutive frame/segment dissimilarity within each modality.
We then fine-tune text-to-music models on music-event curves and swap in video-event curves at inference time.
Thus, we bridge modalities through temporal structure rather than paired supervision while enabling independent timing and semnatic control.
On OES-Pub, MovieGenBench-Music, and AIST++, \ourmodel consistently achieves state-of-the-art results without paired music and video data, outperforming paired-data methods in audio quality, semantic alignment, and temporal synchronization. Direct human evaluation reinforces this.

%% file: 6_appendix.tex
\section*{Appendix overview}

Our appendix consists of:
\begin{enumerate}[itemsep=6pt, topsep=2pt]
    \item Limitations and broader impact.
    \item Implementation details.
    \item Additional quantitative results.
    \item Qualitative results on event curves.
\end{enumerate}

\section{Limitations and broader impact}
\vspace{\subsecmargin}
\label{sec:limitations}

\noindent\textbf{Limitations.}
Existing automatic metrics capture complementary but incomplete aspects of video-to-music generation.
They measure audio quality, coarse semantic alignment, and local synchronization between scene and music changes, but do not fully capture complex temporal structures, macro-level phrasing, or professional scoring quality.
As a result, some extensions of \ourmodel that may improve longer-term musical structure cannot yet be fairly evaluated, and we therefore keep the main system focused on the aspects that can be measured reliably.

\noindent\textbf{Broader impact.}
This work can reduce the effort needed to create time-aligned background music and make video-to-music tools more accessible.
It may be especially useful for commercial systems, since licensed video-music pairs are difficult and expensive to obtain, while licensing text-music data and silent videos is more feasible.
Potential negative impacts include encouraging greater media consumption and screen time, which may raise concerns around digital addiction and social isolation.

\input{code/sch}
\input{code/vibe}
\sloppy
\section{Implementation details}
\vspace{\subsecmargin}
\label{sec:imp}

\noindent\textbf{Architecture.}
We adopt an internal, pretrained T2M model similar to~\cite{novackpresto, bai2025dragon} for fine-tuning.
The audio autoencoder is similar to~\cite{arxiv24_stable_audio, bralios2025re, bai2025dragon, bralios2025learning} and compresses stereo 44.1 kHz waveforms into continuous latents ($d{=}64$) at 12.3 Hz, yielding 394 frames for 32-second clips.
The rectified flow model uses the DiT architecture with $\approx 1$ billion parameters: 16 layers, hidden dimension 2048, and feed-forward dimension 8192.
Text conditioning is implemented via cross-attention layers using Gemma-3B~\cite{team2025gemma} embeddings.

\noindent\textbf{Feature encoder.}
We use MusicFM~\cite{musicfm} and DINOv2-L~\cite{dinov2} to extract music features $\mathbf{f}_m$ and video features $\mathbf{f}_v$ by default. For dance videos, we report results with CoTracker~\cite{cotracker}, while the default DINOv2 setup already exceeds prior AIST++ baselines.
We explore a comprehensive ablation of encoder choices, including an audio autoencoder, AVSiam~\cite{avsiam}, and V-JEPA~\cite{vjepa2} in Section~\ref{sec:abs}.

\noindent\textbf{Training.}
We initialize from a pretrained text-to-music model, adding 2048 parameters to input and project our event curve conditioning, and fine-tune on $\approx 25$k hours of licensed instrumental music-text pairs.
Fine-tuning is lightweight, requiring only 2--4 days on 4--8 A100 GPUs using 32-second clips (192-768 GPU hours).
We use the AdamW optimizer with a learning rate $10^{-4}$ and apply classifier-free guidance with 10\% condition dropout.
Cumulatively, this fine-tuning scheme enables video-to-music generation without paired video-music data or training from scratch, given a pretrained model.

\noindent\textbf{Scene cut hit (SCH).}
To evaluate the temporal consistency between generated music and the underlying video dynamics, we design and use the SCH metric. The intuition is as follows:
When a scene change occurs in the video, well-aligned background music should exhibit a corresponding onset or noticeable rhythmic change.
This principle reflects common filmmaking conventions, where directors often synchronize cuts with musical beats to enhance pacing and emotional impact.
The SCH metric is conceptually similar to beat-matching metrics~\cite {siggraph24_d2m} but describes aperiodic visual events.
Given a video, we first detect its scene cuts using the \texttt{PySceneDetect}, obtaining a set of scene-cut timestamps that serve as temporal anchors.
We then extract the audio track and compute musical onsets using the beat–tracking algorithm provided in \texttt{Librosa}.
While more sophisticated onset detectors are available, beat tracking offers a reproducible, robust, and musically meaningful proxy for rhythmic emphasis and percussion peaks.
For each detected scene cut, we count a \emph{hit} if at least one musical onset falls within a small temporal tolerance (we use $\pm0.1$ seconds). The final SCH score is the ratio of matched cuts to all scene cuts.
The complete procedure is summarized in Algorithm~\ref{alg:scene_hit}, which outlines the pipeline from scene detection to music onset extraction and the computation of the overall hit rate.

\noindent\textbf{VLLM for music prompts.}
To obtain high-level semantic music prompts that reflect the mood/energy/etc. of a video, we adopt the Vibe framework~\cite{vibe} with modern multimodal components.
Given an input video, we first obtain its transcript using the Whisper~\cite{whisper} ASR model, which provides a robust textual representation of the spoken content.
Dialogue and narration often contain emotionally or contextually important cues, making the transcript an essential input to the music–prompt generation process.

In parallel, we sample video frames and extract frame-level visual descriptions using Gemma-4B~\cite{team2025gemma}. These descriptions capture high-level semantic attributes such as environment, actions, character states, and affective cues.
Since individual captions may be redundant, we aggregate them using Gemma-4B, which summarizes the set into a compact, coherent representation of the entire video.

Finally, the LLM conditions jointly on the transcript and the visual summary to produce the final music prompt, which describes mood, instrumentation, intensity, and emotional character. This process encourages prompts that capture multimodal semantics and better align with creators’ intuitive descriptions. The full procedure is provided in Algorithm~\ref{algo:prompt}.

\section{Additional quantitative results.}
\label{sec:additional_results}

\noindent\textbf{Domain-specific visual encoders.}
Our event-curve framework enables performance gains via inference-time encoder selection.
We validate this on dance videos (AIST++), where motion is tightly synchronized to a single object.
Our default configuration with DINOv2 outperforms all the competing methods (BCS: 0.5522, BHS: 0.5748, F1: 0.5750, TD: 17.23).
We hypothesize that CoTracker~\cite{cotracker}, a point-level motion tracker, is more suitable for this task. This is true and improves all metrics (BCS: 0.5818, BHS: 0.6274, F1: 0.5856, TD: 12.24) in \tabref{dance}.
This shows that \ourmodel can be customized at inference by picking a domain-specific encoder.

\begin{table}[t]
\centering
\caption{\textbf{Comparison with a text-only baseline.} We compare \ourmodel against the text-to-music model used for our finetuning, without event-curve conditioning, on OES-Pub.}
\setlength{\tabcolsep}{4pt}
\resizebox{0.4\linewidth}{!}{
\begin{tabular}{lccc}
\toprule
Method & FAD$\downarrow$ & CLAP$\uparrow$ & SCH$\uparrow$ \\
\midrule
Text-only & \textbf{3.63} & 0.23 & 0.35 \\
\ourmodel & 4.95 & 0.23 & \textbf{0.61} \\
\bottomrule
\label{tab:text-only}
\end{tabular}
}
\end{table}

\noindent\textbf{Comparison with the text-only baseline.}
In \tabref{text-only}, we compare \ourmodel with a text-only baseline, \ie the same text-to-music backbone, checkpoint, and predicted music prompts without event-curve conditioning, on OES-Pub.
The results show that event-curve conditioning substantially improves temporal alignment, increasing SCH from 0.35 to \textbf{0.61}, while preserving semantic consistency, as both models achieve the same CLAP score of 0.23.
Although the text-only baseline achieves similar music quality, its much weaker SCH suggests that text guidance alone is insufficient for accurate synchronization.
Overall, these results suggest that event curves improve synchronization by injecting explicit temporal structure, while leaving semantic alignment largely unchanged.

\begin{table}[t]
\centering
\caption{\textbf{Comparison with a large-scale open-source model.} We compare \ourmodel with HunyuanVideo-Foley~\cite{shan2025hunyuanvideo} on OES-Pub.}
\setlength{\tabcolsep}{4pt}
\resizebox{0.48\linewidth}{!}{
\begin{tabular}{lccc}
\toprule
Method & FAD*$\downarrow$ & CLAP$\uparrow$ & SCH$\uparrow$ \\
\midrule
HunyuanVideo-Foley~\cite{shan2025hunyuanvideo} & 15.02 & 0.12 & 0.36 \\
\ourmodel & \textbf{4.95} & \textbf{0.23} & \textbf{0.61} \\
\bottomrule
\end{tabular}
}
\label{tab:hunyuan_compare}
\end{table}

\noindent\textbf{Comparison with a large-scale open-source model.}
In \tabref{hunyuan_compare}, we compare \ourmodel with HunyuanVideo-Foley, a large-scale open-source model designed for general audio generation rather than music generation, which has not been evaluated on video-to-music benchmarks.
Since HunyuanVideo-Foley generates music but also speech and environmental sounds, it performs substantially worse across all three metrics: FAD* (15.02 vs.\ 4.95), CLAP (0.124 vs.\ 0.23), and SCH (0.36 vs.\ 0.61).
However, this comparison should be interpreted with caution, as HunyuanVideo-Foley is not designed for music generation and is therefore not well-suited to this setting or our music-focused evaluation protocol.

\input{supp/fd}

\noindent\textbf{Temporal alignment analysis.} Since the core idea of \ourmodel is based on event curves, we further investigate how event-curve distances relate to low-level temporal alignment quality.
Specifically, since our method is designed to match these event curves, we attempt to quantify and understand the distributional behavior of this tight synchronization.
In \tabref{curve_fd}, we report four variants of Fréchet Distance computed between different event-curve distributions: (1) \textbf{M}, generated to ground-truth music event curves, (2) \textbf{M+V}, concatenated generated music-video event curves and ground-truth music-video event curves. (3) \textbf{M-V}, generated music event curves to video event curves, and (4) \textbf{M$\mid$V}, generated music event curves and ground-truth music event curves, both conditioned on video event curves.
\textbf{M$\mid$V} uses a block partitioned Gaussian to represent the music given video conditional distribution.
The music event curves are extracted by musicfm~\cite{musicfm}, and the video event curves are extreacted by DINOv2~\cite{dinov2}

We compare these additional quantitative results with human judgments of temporal alignment following the protocol used in the main paper.
Interestingly, the human preference results (rightmost column of Table~\ref{tab:curve_fd}) do not correlate strongly with any of the event-curve distances.
Models that achieve low Fréchet distances are not necessarily preferred by human raters, and vice versa.

This mismatch suggests an important insight: event curves may be highly suitable as a generative intermediate representation, but they may not be appropriate as an evaluation metric.
Event curves are good at capturing dense, continuous temporal dynamics that help guide generation, yet humans appear to assess alignment based on sparse, salient temporal moments (e.g., impactful transitions or rhythm–scene coincidences), not global curve similarity.
As a result, metrics that rely purely on curve distributions reward continuous structural alignment, while human perception is more selective and non-uniform over time.

A second contributing factor is that current evaluation datasets (OES-Pub and MovieGenBench-Music) are not specifically curated for temporal synchronization assessment. Many clips lack strong or consistent rhythmic structure, limiting the signal available to curve-based metrics.

In sum, these observations indicate that while \textbf{event curves provide a powerful foundation for generation, but the distributional distances do not necessarily reflect human-perceived temporal alignment quality}. Future benchmarks tailored to temporal synchronization and metrics that account for sparse perceptual saliency may bridge this gap.

\noindent\textbf{Event curve robustness to non-semantic changes.}
To test the robustness of DINOv2 to non-semantic visual changes that may affect event curves, we perform random non-semantic augmentations to 3,450 OES-Pub frames (fps=1), where each frame is randomly perturbed: subpixel translation $\pm 4$ px, rotation $\pm 4^\circ$, brightness step $[0.4, 1.6]\times$, or gamma change $[0.4,1.8]$.
The cosine similarity between DINOv2 features before and after augmentation is $\mu=0.983$ with $\sigma=0.025$, indicating that the video event curves remain robust under these perturbations.

\noindent\textbf{Representation power of event curves.}
Event curves give relative timing information, while text conveys complex semantic guidance.
To further measure the representation power of event curves, we set up a 3-way classification task among cinematic/natural/dance videos using samples from OES-Pub/MovieGenBench/AIST++. Using the video-event curves as input, a 1-layer MLP with a 90\%/10\% train–test split achieved 68.2\% test accuracy, showing that the curves contain meaningful information based on video dynamics.
This shows that video-event curves already capture non-trivial video characteristics, which are further complemented by text prompts in the full model.

\begin{table}[t]
\centering
\caption{\textbf{Video-event curves across different temporal scales.}
We study video-event curves computed from different frame offsets in the similarity calculation using Stable-Audio-ControlNet~\cite{sd_audio_ctrl} on OES-Pub.}
\label{tab:sd_audio_appendix}
\begin{tabular}{lccc}
\toprule
\bf Video-event curve design & \bf FAD*$\downarrow$ & \bf CLAP$\uparrow$ & \bf SCH$\uparrow$ \\
\midrule
No video-event conditioning & 4.86 & \textbf{0.18} & 0.28 \\
60 frames (2\,s) & 3.76 & 0.15 & 0.29 \\
15 frames (500\,ms) & \textbf{3.65} & 0.15 & 0.35 \\
Multiple Offsets (1, 15, 60) & 3.84 & 0.16 & 0.33 \\
\midrule
1 frame (33\,ms, default) & 4.13 & 0.17 & \textbf{0.38} \\
\bottomrule
\end{tabular}
\end{table}

\noindent\textbf{Different temporal dependencies for video-event curves.}
In \tabref{sd_audio_appendix}, we study the effect of different similarity offsets (\ie temporal dependencies) in video-event curve extraction on OES-Pub.
We use Stable-Audio-ControlNet~\cite{sd_audio_ctrl}, which is trained with RMS control rather than music event curves, as a controlled testbed.
Since this setup does not involve music-event-curve offsets during training, we vary only the offset used to compute video-event curves at inference, allowing us to study the effect of video-event temporal scale in isolation.

At 30 fps, offsets 1, 15, and 60 correspond to about 33\,ms, 500\,ms, and 2\,s, respectively.
The multiple-offset variant uses video-event curves computed with offsets 1, 15, and 60 as joint conditioning signals through compositional guidance.
Longer or multiple offsets are feasible and preserve overall audio quality (\ie FAD) and semantic alignment (\ie CLAP).
However, we do not have a metric that properly evaluates longer temporal dependencies across video and music.
Since the default offset-1 design yields the strongest measured synchronization under the available metric (\ie SCH), we keep offset to 1 in the main model and leave longer-term dependencies and multiple-offset conditioning for future work.

\noindent\textbf{Potential extension to macro-level phrasing.}
Our music captioning pipeline currently produces a single prompt for the entire video, which may limit the ability to capture macro-level phrasing changes in music that correspond to different video segments.
However, the same captioning framework could be applied to shorter video chunks, allowing the prompt to change over time and potentially capture more fine-grained music semantic changes.

To test this possibility, we need a shared model that supports both text-to-music (T2M) and video-to-music (V2M) generation, so that chunk-wise T2M outputs can be compared with V2M outputs under the same backbone.
We choose AudioX~\cite{audiox} for this purpose, as it supports V2M and T2M in a single model, which supports previous music chunks as a condition.
In the chunked T2M setting, this allows each new music chunk to be generated from the current chunk-level prompt while conditioning on the previously generated music chunk, making time-varying prompting practical.

We compare whole-video T2M prompting, chunked T2M prompting from 5\,s video segments, and V2M.
We then compute consie similarity with MusicFM~\cite{musicfm} features between T2M output and the V2M output on 5\,s music chunks, then average across chunks within a video.
Chunked prompting is closer to V2M than whole-video prompting (\ie \textbf{0.86} versus \textbf{0.63} for the whole-video prompt),
suggesting that time-varying captions can make T2M generation more like V2M and may help capture macro-level phrasing.
Since there is no direct metric that properly evaluates macro-level phrasing, we treat this as preliminary evidence rather than a main design choice.
We therefore keep full-video prompts in the main model for simplicity, and leave systematic chunk-wise prompting for future work.

\begin{figure}[t]
    \centering
    \includegraphics[width=0.9\linewidth]{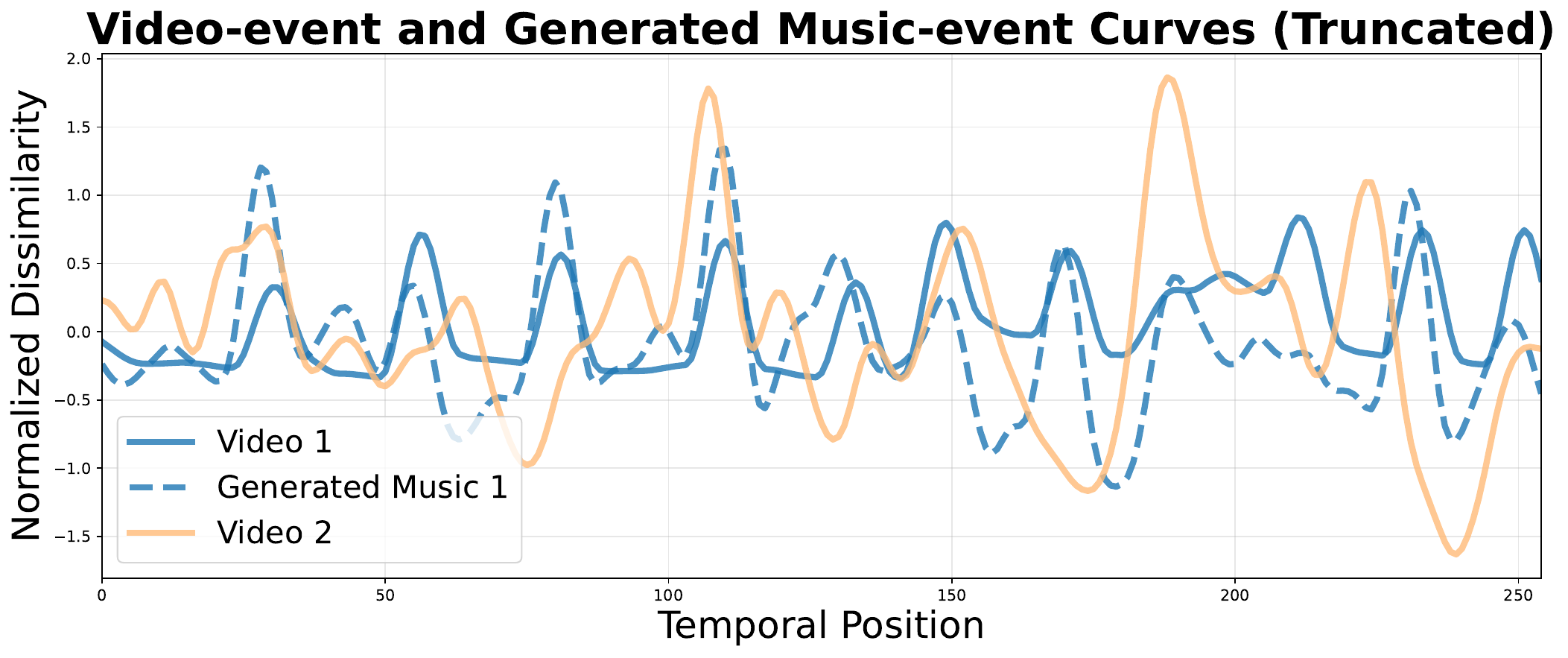}
    \caption{\textbf{Example event curves with different temporal dynamics.}
    The blue solid curve corresponds to a video with frequent scene cuts, while the orange curve corresponds to a video with slower visual motion, showing distinct temporal structures. This supports our design choice of using event curves to represent relative timing, while text provides complementary semantic guidance.}
    \label{fig:two_visual_one_music}
\end{figure}

\section{Qualitative results on event curves}

\noindent\textbf{Event curves over diverse dynamics.}
\ourmodel is conditioned on both event curves and text. Event curves with normalization provide relative temporal structure, aiming to represent \emph{when} changes happen, while text conveys complementary information such as semantic content, energy, and mood.

To validate this behavior, \figref{two_visual_one_music} shows two videos with distinct visual dynamics together with an example generated music event curve. The blue solid curve corresponds to a video with frequent scene cuts (\ie Good Sample, (Sora2) 20251120\_0147\_01kaengrpqetrrwv7jyycmnkzn.mp4), while the orange curve corresponds to a video dominated by slower camera motion, such as pans (\ie Random Sample, (MovieGen) 148.mp4, in demo).
Although both visual signals are normalized, their temporal patterns remain clearly different: the scene-cut video exhibits sharp, sparse peaks, whereas the slow-pan video varies more smoothly over time. This example shows that normalization does not erase temporal structure; instead, it preserves relative changes that are useful for synchronization.

\noindent\textbf{Fine-grained temporal precision.}
Although event curves are extracted using temporal smoothing, the resulting peaks (\ie video 1 in \figref{two_visual_one_music}) still retain precise temporal localization. We use a Hann window, whose central lobe preserves peak locations while suppressing high-frequency noise. In equivalent bandwidth, our smoothing corresponds to roughly 21 samples at 30 fps, or about a 700\,ms window. In practice, this smoothing improves robustness without preventing accurate localization of salient events, which is consistent with our use of a $\pm 100$\,ms tolerance in Scene Cut Hit (SCH).

%% file: code/sch.tex
\begin{algorithm}[!b]
\caption{Implementation details of scene cut hit.}
\label{alg:scene_hit}
\fontsize{7}{8}\selectfont
\begin{lstlisting}[
    style=pythonstyle,
    language=Python,
    basicstyle=\ttfamily\fontsize{7}{8}\selectfont,
    aboveskip=1pt,
    belowskip=1pt,
    xleftmargin=2pt,
    framexleftmargin=2pt,
    columns=fullflexible,
    keepspaces=true
]
import torchaudio
import librosa
from scenedetect import detect, AdaptiveDetector

def process_video(video_path, tolerance=0.1):
    # Detect scenes
    scene_list = detect(video_path, AdaptiveDetector())
    scene_cut_times = []
    for i, (start, end) in enumerate(scene_list):
        if i < len(scene_list) - 1:
            scene_cut_times.append(end.get_seconds())

    if len(scene_cut_times) == 0:
        return None

    # Load audio
    x, sr = torchaudio.load(video_path)
    x = x.mean(0).numpy()

    # Beat/onset extraction
    tempo, beats = librosa.beat.beat_track(y=x, sr=sr)
    onsets = librosa.frames_to_time(beats, sr=sr)

    # Hit-rate computation
    hits = 0
    for c in scene_cut_times:
        for o in onsets:
            if abs(o - c) <= tolerance:
                hits += 1
                break

    return hits / len(scene_cut_times)
\end{lstlisting}
\end{algorithm}

%% file: code/vibe.tex
\begin{algorithm}[h]
\caption{\textbf{Music prompt generation from video}~\cite{vibe}}
\begin{algorithmic}[1]
\Require Video $V$, Automatic Speech Recognition $\mathcal{A}$, Vision-Language Model $\mathcal{L}_v$, Large Language Model $\mathcal{L}$
\State Extract transcript: $T \leftarrow \mathcal{A}(V)$
\State Initialize caption set: $C \leftarrow \varnothing$
\For{each sampled frame $f_i$ from $V$}
    \State Obtain visual description: $C_i \leftarrow \mathcal{L}_v(f_i)$
    \State $C \leftarrow C \cup \{C_i\}$
\EndFor
\State Generate visual summary: $S \leftarrow \mathcal{L}(C)$
\State Generate music prompt: $P \leftarrow \mathcal{L}(T, S)$
\State \Return $P$
\end{algorithmic}
\label{algo:prompt}
\end{algorithm}

%% file: supp/fd.tex
\begin{table*}[!t]
\caption{\textbf{Event-curve Fréchet distance comparison.}
M evaluates generated vs.\ ground-truth music event curves.
M+V compares concatenated music--video event curves.
M-V compares the distribution of generated music and ground-truth video.
M$\mid$V evaluates conditional music event distributions given video.
We additionally report human preference for temporal alignment. $^*$ denotes statistical significance.}
\centering
\resizebox{0.95\linewidth}{!}{
\begin{tabular}{l|cccc|cccc|c}
\toprule
& \multicolumn{4}{c|}{\textbf{OES-Pub}} &
\multicolumn{4}{c|}{\textbf{MovieGenBench-Music}} &
\textbf{Human Eval} \\
\cmidrule(lr){2-5} \cmidrule(lr){6-9} \cmidrule(lr){10-10}
\textbf{Method} &
\textbf{M} &
\textbf{M+V} &
\textbf{M-V} &
$\textbf{M}\mid\textbf{V}$ &
\textbf{M} &
\textbf{M+V} &
\textbf{M-V} &
$\textbf{M}\mid\textbf{V}$ &
\textbf{Temporal Align.} \\
\midrule
M$^{2}$UGen~\cite{arxiv24_mumu} & 6.29 & 6.96 & 18.26 & 1.33 & 1.96 & 1.99 & 24.76 & 1.52 & Ours wins$^*$ \\
GVMGen~\cite{aaai25_gvmgen}        & 3.45 & 4.04 & 10.10 & 0.48 & 1.31 & 1.34 & 20.09 & 1.02 & Ours wins$^*$ \\
MTCV2M~\cite{acmmm25_controllable} & 4.33 & 4.53 & 10.90 & 0.92 & 1.53 & 1.75 & 21.03 & 0.98 & Ours wins$^*$ \\
VidMuse~\cite{cvpr25_vidmuse}   & 3.25 & 3.86 & 12.40 & 0.79 & 4.27 & 4.29 & 40.89 & 4.02 & Ours wins \\
AudioX~\cite{audiox}            & 4.13 & 4.63 &  1.75 & 1.93 & 1.52 & 1.54 & 19.70 & 1.18 & Ours wins$^*$ \\
\midrule
SONIQUE~\cite{icassp25_sonique} & 3.13 & 3.75 &  8.50 & 0.15 &11.10 &11.12 &  6.33 & 9.75 & Ours wins$^*$ \\
\midrule
\ourmodel (Ours)                & 5.34 & 5.87 &  1.20 & 3.09 & 3.27 & 3.29 & 27.90 & 1.75 & N/A \\
\bottomrule
\end{tabular}}
\label{tab:curve_fd}
\end{table*}

%% file: egbib.bib
@String{tmm = "IEEE Transactions on Multimedia (TMM)"}

@String{tip = "TIP"}

@String{cvpr = "CVPR"}

@String{acmmm = "ACM MM"}

@String{nips = "NeurIPS"}

@String{aaai = "AAAI"}

@String{tmlr = "TMLR"}

@String{ismir = "ISMIR"}

@String{eccv = "ECCV"}

@String{eccvw = "ECCVW"}

@String{iclr = "ICLR"}

@String{iccv = "ICCV"}

@String{icml = "ICML"}

@String{wacv = "WACV"}

@String{icassp = "ICASSP"}

@String{arxiv = "arXiv Preprint"}

@String{taslp = "TASLP"}

@inproceedings{gu2026vmsp,
  title={VMSP: Video-to-Music Generation with Two-Stage Alignment and Synthesis},
  author={Gu, Xin and Jiang, Wei and Jiang, Yujian and Su, Zhibin and Yan, Ming},
  booktitle=icassp,
  year={2026},
}

@inproceedings{tian2026audio_omni,
  title={Audio-Omni: Extending Multi-modal Understanding to Versatile Audio Generation and Editing},
  author={Tian, Zeyue and Yang, Binxin and Liu, Zhaoyang and Zhang, Jiexuan and Yuan, Ruibin and Yin, Hubery and Chen, Qifeng and Li, Chen and Lv, Jing and Xue, Wei and others},
  booktitle={SIGGRAPH} ,
  year={2026}
}

@inproceedings{whisper,
  title={Robust speech recognition via large-scale weak supervision},
  author={Radford, Alec and Kim, Jong Wook and Xu, Tao and Brockman, Greg and McLeavey, Christine and Sutskever, Ilya},
  booktitle=icml,
  year={2023},
}

@inproceedings{cvpr25_mmaudio,
  title={MMAudio: Taming Multimodal Joint Training for High-Quality Video-to-Audio Synthesis},
  author={Cheng, Ho Kei and Ishii, Masato and Hayakawa, Akio and Shibuya, Takashi and Schwing, Alexander and Mitsufuji, Yuki},
  booktitle=cvpr,
  year={2025}
}

@inproceedings{su2024vision,
  title={From vision to audio and beyond: A unified model for audio-visual representation and generation},
  author={Su, Kun and Liu, Xiulong and Shlizerman, Eli},
  booktitle=icml,
  year={2024}
}

@article{tong2025videoechoedmusicsemantic,
      title={Video Echoed in Music: Semantic, Temporal, and Rhythmic Alignment for Video-to-Music Generation}, 
      author={Xinyi Tong and Yiran Zh and Jishang Chen and Chunru Zhan and Tianle Wang and Sirui Zhang and Nian Liu and Tiezheng Ge and Duo Xu and Xin Jin and Feng Yu and Song-Chun Zhu},
      year={2025},
      journal=arXiv
}

@article{wang2025kling,
  title={{Kling-Foley}: Multimodal Diffusion Transformer for High-Quality Video-to-Audio Generation},
  author={Wang, Jun and Zeng, Xijuan and Qiang, Chunyu and Chen, Ruilong and Wang, Shiyao and Wang, Le and Zhou, Wangjing and Cai, Pengfei and Zhao, Jiahui and Li, Nan and others},
  journal=arxiv,
  year={2025}
}

@inproceedings{liu2024tell,
  title={Tell what you hear from what you see-video to audio generation through text},
  author={Liu, Xiulong and Su, Kun and Shlizerman, Eli},
  booktitle=nips,
  year={2024}
}

@article{llama3,
  title={The Llama 3 Herd of Models},
  author={Grattafiori, Aaron and Dubey, Abhimanyu and Jauhri, Abhinav and Pandey, Abhinav and Kadian, Abhishek and Al-Dahle, Ahmad and Letman, Aiesha and Mathur, Akhil and Schelten, Alan and Vaughan, Alex and others},
  journal=arxiv,
  year={2024}
}

@article{qwen3,
  title={Qwen3 technical report},
  author={Yang, An and Li, Anfeng and Yang, Baosong and Zhang, Beichen and Hui, Binyuan and Zheng, Bo and Yu, Bowen and Gao, Chang and Huang, Chengen and Lv, Chenxu and others},
  journal=arxiv,
  year={2025}
}

@inproceedings{cvpr25_harmonyset,
  title={Harmonyset: A comprehensive dataset for understanding video-music semantic alignment and temporal synchronization},
  author={Zhou, Zitang and Mei, Ke and Lu, Yu and Wang, Tianyi and Rao, Fengyun},
  booktitle=cvpr,
  year={2025}
}

@inproceedings{cotracker,
  title={Cotracker: It is better to track together},
  author={Karaev, Nikita and Rocco, Ignacio and Graham, Benjamin and Neverova, Natalia and Vedaldi, Andrea and Rupprecht, Christian},
  booktitle=eccv,
  year={2024},
}

@article{vjepa2,
  title={{V-JEPA} 2: Self-supervised video models enable understanding, prediction and planning},
  author={Assran, Mido and Bardes, Adrien and Fan, David and Garrido, Quentin and Howes, Russell and Muckley, Matthew and Rizvi, Ammar and Roberts, Claire and Sinha, Koustuv and Zholus, Artem and others},
  journal=arxiv,
  year={2025}
}

@article{dinov2,
  title={{DINOv2}: Learning robust visual features without supervision},
  author={Oquab, Maxime and Darcet, Timoth{\'e}e and Moutakanni, Th{\'e}o and Vo, Huy and Szafraniec, Marc and Khalidov, Vasil and Fernandez, Pierre and Haziza, Daniel and Massa, Francisco and El-Nouby, Alaaeldin and others},
  journal=arxiv,
  year={2023}
}

@inproceedings{musicfm,
  title={A foundation model for music informatics},
  author={Won, Minz and Hung, Yun-Ning and Le, Duc},
  booktitle=icassp,
  year={2024},
}

@inproceedings{avsiam,
  title={Siamese vision transformers are scalable audio-visual learners},
  author={Lin, Yan-Bo and Bertasius, Gedas},
  booktitle=eccv,
  year={2024},
}

@inproceedings{siggraph23_motion2dance,
  title={Motion to dance music generation using latent diffusion model},
  author={Tan, Vanessa and Nam, Junghyun and Nam, Juhan and Noh, Junyong},
  booktitle={SIGGRAPH Asia},
  year={2023}
}

@article{arxiv23_mu,
  title={{M$^{2}$ UGen}: Multi-modal Music Understanding and Generation with the Power of Large Language Models},
  author={Liu, Shansong and Hussain, Atin Sakkeer and Wu, Qilong and Sun, Chenshuo and Shan, Ying},
  journal=arxiv,
  year={2023}
}

@article{arxiv24_mumu,
  title={{MuMu-LLaMA}: Multi-modal Music Understanding and Generation via Large Language Models},
  author={Liu, Shansong and Hussain, Atin Sakkeer and Wu, Qilong and Sun, Chenshuo and Shan, Ying},
  journal=arxiv,
  year={2024}
}

@incollection{menell2022musiclicensing,
  author    = {Peter S. Menell and Suzanne Scotchmer},
  title     = {Music Licensing},
  booktitle = {Intellectual Property Licensing and Transactions: Theory and Practice},
  publisher = {Cambridge University Press},
  year      = {2022},
  pages     = {493--522},
  doi       = {10.1017/9781009049436.017},
  url       = {https://doi.org/10.1017/9781009049436.017}
}

@incollection{rosario2024sync,
  author    = {Lita Rosario-Richardson and John Simson},
  title     = {Synchronization Licensing in the New Music Digital Ecosystem},
  booktitle = {The Oxford Handbook of Music Law and Policy},
  publisher = {Oxford University Press},
  year      = {2024},
  doi       = {10.1093/oxfordhb/9780190872243.013.40},
  url       = {https://doi.org/10.1093/oxfordhb/9780190872243.013.40}
}

@article{yinghao2024foundation,
  title={Foundation models for music: A survey},
  author={Yinghao, M and Anders, {\O} and Anton, R and Del, S Bleiz MacSen and Charalampos, S and Chris, D},
  journal={arXiv preprint arXiv:2408.14340},
  volume={3},
  year={2024}
}

@article{song,
  title={The song describer dataset: a corpus of audio captions for music-and-language evaluation},
  author={Manco, Ilaria and Weck, Benno and Doh, Seungheon and Won, Minz and Zhang, Yixiao and Bogdanov, Dmitry and Wu, Yusong and Chen, Ke and Tovstogan, Philip and Benetos, Emmanouil and others},
  journal=arxiv,
  year={2023}
}

@inproceedings{aist++,
  title={{AI} choreographer: Music conditioned 3d dance generation with {AIST++}},
  author={Li, Ruilong and Yang, Shan and Ross, David A and Kanazawa, Angjoo},
  booktitle=iccv,
  year={2021}
}

@inproceedings{aist,
  title={{AIST} Dance Video Database: Multi-Genre, Multi-Dancer, and Multi-Camera Database for Dance Information Processing.},
  author={Tsuchida, Shuhei and Fukayama, Satoru and Hamasaki, Masahiro and Goto, Masataka},
  booktitle={ISMIR},
  year={2019}
}

@article{moviegen,
  title={{Movie Gen}: A cast of media foundation models},
  author={Polyak, Adam and Zohar, Amit and Brown, Andrew and Tjandra, Andros and Sinha, Animesh and Lee, Ann and Vyas, Apoorv and Shi, Bowen and Ma, Chih-Yao and Chuang, Ching-Yao and others},
  journal=arxiv,
  year={2024}
}

@inproceedings{vibe,
  title={``It’s more of a vibe I’m going for'': Designing Text-to-Music Generation Interfaces for Video Creators},
  author={Hammad, Noor and Fraser, C Ailie and Harpstead, Erik and Hammer, Jessica and Dontcheva, Mira},
  booktitle={DIS},
  year={2025}
}

@inproceedings{dac,
  title={High-fidelity audio compression with improved {RVQGAN}},
  author={Kumar, Rithesh and Seetharaman, Prem and Luebs, Alejandro and Kumar, Ishaan and Kumar, Kundan},
  booktitle=nips,
  year={2023}
}

@article{encodec,
  title={High fidelity neural audio compression},
  author={D{\'e}fossez, Alexandre and Copet, Jade and Synnaeve, Gabriel and Adi, Yossi},
  journal=arxiv,
  year={2022}
}

@article{soundstream,
  title={Soundstream: An end-to-end neural audio codec},
  author={Zeghidour, Neil and Luebs, Alejandro and Omran, Ahmed and Skoglund, Jan and Tagliasacchi, Marco},
  journal=taslp,
  year={2021},
}

@article{rowles2025foley,
  title={Foley Control: Aligning a Frozen Latent Text-to-Audio Model to Video},
  author={Rowles, Ciara and Jampani, Varun and Donn{\'e}, Simon and Vainer, Shimon and Parker, Julian and Evans, Zach},
  journal=arxiv,
  year={2025}
}

@article{arxiv25_zs_musiscene,
  title={{MusiScene}: Leveraging MU-LLaMA for Scene Imagination and Enhanced Video Background Music Generation},
  author={Izzati, Fathinah and Li, Xinyue and Wu, Yuxuan and Xia, Gus},
  journal=arxiv,
  year={2025}
}

@article{arxiv24_zs_m2m,
  title={{M2M-Gen}: A multimodal framework for automated background music generation in japanese manga using large language models},
  author={Sharma, Megha and Haseeb, Muhammad Taimoor and Xia, Gus and Tsuruoka, Yoshimasa},
  journal=arxiv,
  year={2024}
}

@inproceedings{aigc25_zs_mozart,
  title={Mozart’s Touch: a lightweight multimodal music generation framework based on pre-trained large models},
  author={Li, Jiajun and Xu, Tianze and Chen, Xuesong and Yao, Xinrui and Han, Jingchou and Liu, Shuchang},
  booktitle={AIGC},
  year={2025},
}

@article{arxiv24_zs_multimodal,
  title={Multimodal music generation with explicit bridges and retrieval augmentation},
  author={Wang, Baisen and Zhuo, Le and Wang, Zhaokai and Bao, Chenxi and Chengjing, Wu and Nie, Xuecheng and Dai, Jiao and Han, Jizhong and Liao, Yue and Liu, Si},
  journal=arxiv,
  year={2024}
}

@inproceedings{icassp25_sonique,
  title={{SONIQUE}: Video background music generation using unpaired audio-visual data},
  author={Zhang, Liqian and Fuentes, Magdalena},
  booktitle=icassp,
  year={2025},
}

@article{arxiv25_yue,
  title={Yue: Scaling open foundation models for long-form music generation},
  author={Yuan, Ruibin and Lin, Hanfeng and Guo, Shuyue and Zhang, Ge and Pan, Jiahao and Zang, Yongyi and Liu, Haohe and Liang, Yiming and Ma, Wenye and Du, Xingjian and others},
  journal=arxiv,
  year={2025}
}

@article{audiox,
  title={AudioX: Diffusion transformer for anything-to-audio generation},
  author={Tian, Zeyue and Jin, Yizhu and Liu, Zhaoyang and Yuan, Ruibin and Tan, Xu and Chen, Qifeng and Xue, Wei and Guo, Yike},
  journal=arxiv,
  year={2025}
}

@inproceedings{nips25_thinksound,
  title={{ThinkSound}: Chain-of-Thought Reasoning in Multimodal Large Language Models for Audio Generation and Editing},
  author={Liu, Huadai and Wang, Jialei and Luo, Kaicheng and Wang, Wen and Chen, Qian and Zhao, Zhou and Xue, Wei},
  booktitle=nips,
  year={2025}
}

@inproceedings{icml25_ditto,
  title={{DITTO}: Diffusion inference-time T-optimization for music generation},
  author={Novack, Zachary and McAuley, Julian and Berg-Kirkpatrick, Taylor and Bryan, Nicholas J.},
  booktitle=icml,
  year={2025}
}

@inproceedings{cvpr25_vintage,
  title={{VinTAGe}: Joint video and text conditioning for holistic audio generation},
  author={Kushwaha, Saksham Singh and Tian, Yapeng},
  booktitle=cvpr,
  year={2025}
}

@inproceedings{cvpr25_multifo,
  title={Video-guided foley sound generation with multimodal controls},
  author={Chen, Ziyang and Seetharaman, Prem and Russell, Bryan and Nieto, Oriol and Bourgin, David and Owens, Andrew and Salamon, Justin},
  booktitle=cvpr,
  year={2025}
}

@inproceedings{cvpr24_melfusion,
  title={{MeLFusion}: Synthesizing music from image and language cues using diffusion models},
  author={Chowdhury, Sanjoy and Nag, Sayan and Joseph, KJ and Srinivasan, Balaji Vasan and Manocha, Dinesh},
  booktitle=cvpr,
  year={2024}
}

@inproceedings{cvpr24_diffbgm,
  title={{Diff-BGM}: A diffusion model for video background music generation},
  author={Li, Sizhe and Qin, Yiming and Zheng, Minghang and Jin, Xin and Liu, Yang},
  booktitle=cvpr,
  year={2024}
}

@article{arxiv24_muvi,
  title={{MuVi}: Video-to-music generation with semantic alignment and rhythmic synchronization},
  author={Li, Ruiqi and Zheng, Siqi and Cheng, Xize and Zhang, Ziang and Ji, Shengpeng and Zhao, Zhou},
  journal=arxiv,
  year={2024}
}

@article{arxiv24_vidmusician,
  title={{VidMusician}: Video-to-music generation with semantic-rhythmic alignment via hierarchical visual features},
  author={Li, Sifei and Yang, Binxin and Yin, Chunji and Sun, Chong and Zhang, Yuxin and Dong, Weiming and Li, Chen},
  journal=arxiv,
  year={2024}
}

@article{tian2025_xmusic,
  title={{XMusic}: Towards a generalized and controllable symbolic music generation framework},
  author={Tian, Sida and Zhang, Can and Yuan, Wei and Tan, Wei and Zhu, Wenjie},
  journal={TMM},
  year={2025},
}

@inproceedings{acmmm25_controllable,
  title={Controllable Video-to-Music Generation with Multiple Time-Varying Conditions},
  author={Wu, Junxian and You, Weitao and Zuo, Heda and Zhang, Dengming and Chen, Pei and Sun, Lingyun},
  booktitle=acmmm,
  year={2025}
}

@inproceedings{aaai25_gvmgen,
  title={{GVMGen}: A general video-to-music generation model with hierarchical attentions},
  author={Zuo, Heda and You, Weitao and Wu, Junxian and Ren, Shihong and Chen, Pei and Zhou, Mingxu and Lu, Yujia and Sun, Lingyun},
  booktitle=aaai,
  year={2025}
}

@inproceedings{ismir25_OES,
  title={Video-Guided Text-to-Music Generation Using Public Domain Movie Collections},
  author={Kim, Haven and Novack, Zachary and Xu, Weihan and McAuley, Julian and Dong, Hao-Wen},
  booktitle={ISMIR},
  year={2025}
}

@article{arxiv25_extending,
  title={Extending Visual Dynamics for Video-to-Music Generation},
  author={Liu, Xiaohao and Tu, Teng and Ma, Yunshan and Chua, Tat-Seng},
  journal=arxiv,
  year={2025}
}

@inproceedings{cvpr25_vidmuse,
  title={{VidMuse}: A simple video-to-music generation framework with long-short-term modeling},
  author={Tian, Zeyue and Liu, Zhaoyang and Yuan, Ruibin and Pan, Jiahao and Liu, Qifeng and Tan, Xu and Chen, Qifeng and Xue, Wei and Guo, Yike},
  booktitle=cvpr,
  year={2025}
}

@inproceedings{cvpr25_filmcomposer,
  title={{FilmComposer}: LLM-Driven Music Production for Silent Film Clips},
  author={Xie, Zhifeng and He, Qile and Zhu, Youjia and He, Qiwei and Li, Mengtian},
  booktitle=cvpr,
  year={2025}
}

@inproceedings{lin2025vmas,
  title={{VMAS}: Video-to-music generation via semantic alignment in web music videos},
  author={Lin, Yan-Bo and Tian, Yu and Yang, Linjie and Bertasius, Gedas and Wang, Heng},
  booktitle=wacv,
  year={2025},
}

@inproceedings{cvpr25_enhancing,
  title={Enhancing Dance-to-Music Generation via Negative Conditioning Latent Diffusion Model},
  author={Sun, Changchang and Liu, Gaowen and Fleming, Charles and Yan, Yan},
  booktitle=cvpr,
  year={2025}
}

@inproceedings{siggraph24_d2m,
  title={Dance-to-music generation with encoder-based textual inversion},
  author={Li, Sifei and Dong, Weiming and Zhang, Yuxin and Tang, Fan and Ma, Chongyang and Deussen, Oliver and Lee, Tong-Yee and Xu, Changsheng},
  booktitle={SIGGRAPH Asia},
  year={2024}
}

@article{arxiv24_flux_music,
  title={Flux that plays music},
  author={Fei, Zhengcong and Fan, Mingyuan and Yu, Changqian and Huang, Junshi},
  journal=arxiv,
  year={2024}
}

@inproceedings{icml25_song_gen,
  title={{SongGen}: A single stage auto-regressive transformer for text-to-song generation},
  author={Liu, Zihan and Ding, Shuangrui and Zhang, Zhixiong and Dong, Xiaoyi and Zhang, Pan and Zang, Yuhang and Cao, Yuhang and Lin, Dahua and Wang, Jiaqi},
  booktitle=icml,
  year={2025}
}

@article{arxiv25_ace_step,
  title={{ACE-Step}: A step towards music generation foundation model},
  author={Gong, Junmin and Zhao, Sean and Wang, Sen and Xu, Shengyuan and Guo, Joe},
  journal=arxiv,
  year={2025}
}

@article{tmlr25_t2m,
  title={Auto-Regressive vs Flow-Matching: a Comparative Study of Modeling Paradigms for Text-to-Music Generation},
  author={Tal, Or and Kreuk, Felix and Adi, Yossi},
  journal=tmlr,
  year={2025}
}

@article{arxiv25_fast_t2a,
  title={Fast Text-to-Audio Generation with Adversarial Post-Training},
  author={Novack, Zachary and Evans, Zach and Zukowski, Zack and Taylor, Josiah and Carr, CJ and Parker, Julian and Al-Sinan, Adnan and Iodice, Gian Marco and McAuley, Julian and Berg-Kirkpatrick, Taylor and others},
  journal=arxiv,
  year={2025}
}

@inproceedings{arxiv24_stable_audio,
  title={Stable audio open},
  author={Evans, Zach and Parker, Julian D and Carr, CJ and Zukowski, Zack and Taylor, Josiah and Pons, Jordi},
  booktitle=icassp,
  year={2025},
}

@article{cfg,
  title={Classifier-free diffusion guidance},
  author={Ho, Jonathan and Salimans, Tim},
  journal=arxiv,
  year={2022}
}

@book{wegele2014max,
  title={Max Steiner: Composing, Casablanca, and the Golden Age of Film Music},
  author={Wegele, Peter},
  year={2014},
  publisher={Bloomsbury Publishing PLC}
}

@inproceedings{arxiv24_soundify,
  title     = {Images that Sound: Composing Images and Sounds on a Single Canvas},
  author    = {Chen, Ziyang and Geng, Daniel and Owens, Andrew},
  booktitle=nips,
  year      = {2024},
}

@inproceedings{icml23_long_video_music,
  title={Long-term rhythmic video soundtracker},
  author={Yu, Jiashuo and Wang, Yaohui and Chen, Xinyuan and Sun, Xiao and Qiao, Yu},
  booktitle=icml,
  year={2023},
}

@inproceedings{iclr23_discrete_music,
  title={Discrete contrastive diffusion for cross-modal music and image generation},
  author={Zhu, Ye and Wu, Yu and Olszewski, Kyle and Ren, Jian and Tulyakov, Sergey and Yan, Yan},
  booktitle=iclr,
  year={2023}
}

@inproceedings{eccv22_dnace2music,
  title={Quantized {GAN} for complex music generation from dance videos},
  author={Zhu, Ye and Olszewski, Kyle and Wu, Yu and Achlioptas, Panos and Chai, Menglei and Yan, Yan and Tulyakov, Sergey},
  booktitle=eccv,
  year={2022},
}

@article{fad,
  title={Fr$\backslash$'echet Audio Distance: A Metric for Evaluating Music Enhancement Algorithms},
  author={Kilgour, Kevin and Zuluaga, Mauricio and Roblek, Dominik and Sharifi, Matthew},
  journal=arxiv,
  year={2018}
}

@inproceedings{vggish,
  title={CNN architectures for large-scale audio classification},
  author={Hershey, Shawn and Chaudhuri, Sourish and Ellis, Daniel PW and Gemmeke, Jort F and Jansen, Aren and Moore, R Channing and Plakal, Manoj and Platt, Devin and Saurous, Rif A and Seybold, Bryan and others},
  booktitle=icassp,
  year={2017},
}

@article{arxiv24_fastaudio,
  title={Fast Timing-Conditioned Latent Audio Diffusion},
  author={Evans, Zach and Carr, CJ and Taylor, Josiah and Hawley, Scott H and Pons, Jordi},
  journal=arxiv,
  year={2024}
}

@inproceedings{iclr24_MAGNET,
  title={Masked Audio Generation using a Single Non-Autoregressive Transformer},
  author={Ziv, Alon and Gat, Itai and Lan, Gael Le and Remez, Tal and Kreuk, Felix and D{\'e}fossez, Alexandre and Copet, Jade and Synnaeve, Gabriel and Adi, Yossi},
  booktitle=iclr,
  year={2024}
}

@article{arxiv23_suitable_video2music,
  title={Video2Music: Suitable Music Generation from Videos using an Affective Multimodal Transformer model},
  author={Kang, Jaeyong and Poria, Soujanya and Herremans, Dorien},
  journal=arxiv,
  year={2023}
}

@inproceedings{aaai24_v2meow,
  title={{V2Meow}: Meowing to the Visual Beat via Music Generation},
  author={Su, Kun and Li, Judith Yue and Huang, Qingqing and Kuzmin, Dima and Lee, Joonseok and Donahue, Chris and Sha, Fei and Jansen, Aren and Wang, Yu and Verzetti, Mauro and others},
  booktitle=aaai,
  year={2024}
}

@inproceedings{iccv23_v2music,
  title={Video background music generation: Dataset, method and evaluation},
  author={Zhuo, Le and Wang, Zhaokai and Wang, Baisen and Liao, Yue and Bao, Chenxi and Peng, Stanley and Han, Songhao and Zhang, Aixi and Fang, Fei and Liu, Si},
  booktitle=iccv,
  year={2023}
}

@inproceedings{acmmm21_cmt,
  title={Video background music generation with controllable music transformer},
  author={Di, Shangzhe and Jiang, Zeren and Liu, Si and Wang, Zhaokai and Zhu, Leyan and He, Zexin and Liu, Hongming and Yan, Shuicheng},
  booktitle=acmmm,
  year={2021}
}

@inproceedings{shlizerman2018audio,
  title={Audio to body dynamics},
  author={Shlizerman, Eli and Dery, Lucio and Schoen, Hayden and Kemelmacher-Shlizerman, Ira},
  booktitle=cvpr,
  year={2018}
}

@article{zhuang2022music2dance,
  title={Music2dance: Dancenet for music-driven dance generation},
  author={Zhuang, Wenlin and Wang, Congyi and Chai, Jinxiang and Wang, Yangang and Shao, Ming and Xia, Siyu},
  journal={TOMM},
  year={2022},
}

@inproceedings{nips19_dance2music,
  title={Dancing to music},
  author={Lee, Hsin-Ying and Yang, Xiaodong and Liu, Ming-Yu and Wang, Ting-Chun and Lu, Yu-Ding and Yang, Ming-Hsuan and Kautz, Jan},
  booktitle=nips,
  year={2019}
}

@inproceedings{cvpr23_v2a,
  title={Conditional Generation of Audio from Video via Foley Analogies},
  author={Du, Yuexi and Chen, Ziyang and Salamon, Justin and Russell, Bryan and Owens, Andrew},
  booktitle=cvpr,
  year={2023}
}

@article{arxiv23_codi2,
  title={Codi-2: In-context, interleaved, and interactive any-to-any generation},
  author={Tang, Zineng and Yang, Ziyi and Khademi, Mahmoud and Liu, Yang and Zhu, Chenguang and Bansal, Mohit},
  journal=arxiv,
  year={2023}
}

@inproceedings{nips23_codi,
  title={Any-to-any generation via composable diffusion},
  author={Tang, Zineng and Yang, Ziyi and Zhu, Chenguang and Zeng, Michael and Bansal, Mohit},
  booktitle=nips,
  year={2023}
}

@article{arxiv23_mustango,
  title={Mustango: Toward controllable text-to-music generation},
  author={Melechovsky, Jan and Guo, Zixun and Ghosal, Deepanway and Majumder, Navonil and Herremans, Dorien and Poria, Soujanya},
  journal=arxiv,
  year={2023}
}

@inproceedings{acmmm23_tango,
  title={Text-to-Audio Generation using Instruction Guided Latent Diffusion Model},
  author={Ghosal, Deepanway and Majumder, Navonil and Mehrish, Ambuj and Poria, Soujanya},
  booktitle=acmmm,
  year={2023}
}

@inproceedings{icml23_make_an_audio,
  title={Make-an-audio: Text-to-audio generation with prompt-enhanced diffusion models},
  author={Huang, Rongjie and Huang, Jiawei and Yang, Dongchao and Ren, Yi and Liu, Luping and Li, Mingze and Ye, Zhenhui and Liu, Jinglin and Yin, Xiang and Zhao, Zhou},
  booktitle=icml,
  year={2023}
}

@inproceedings{audiogen,
  title={Audiogen: Textually guided audio generation},
  author={Kreuk, Felix and Synnaeve, Gabriel and Polyak, Adam and Singer, Uriel and D{\'e}fossez, Alexandre and Copet, Jade and Parikh, Devi and Taigman, Yaniv and Adi, Yossi},
  booktitle=iclr,
  year={2023}
}

@inproceedings{musicgen,
  title={Simple and controllable music generation},
  author={Copet, Jade and Kreuk, Felix and Gat, Itai and Remez, Tal and Kant, David and Synnaeve, Gabriel and Adi, Yossi and D{\'e}fossez, Alexandre},
  booktitle=nips,
  year=2023
}

@inproceedings{clap,
  title={Large-scale contrastive language-audio pretraining with feature fusion and keyword-to-caption augmentation},
  author={Wu, Yusong and Chen, Ke and Zhang, Tianyu and Hui, Yuchen and Berg-Kirkpatrick, Taylor and Dubnov, Shlomo},
  booktitle=icassp,
  year={2023},
}

@article{taslp23_diffsound,
  title={Diffsound: Discrete diffusion model for text-to-sound generation},
  author={Yang, Dongchao and Yu, Jianwei and Wang, Helin and Wang, Wen and Weng, Chao and Zou, Yuexian and Yu, Dong},
  journal=taslp,
  year={2023},
}

@article{arxiv23_Mousais,
  title={{Mo\^{u}sai}: Text-to-music generation with long-context latent diffusion},
  author={Schneider, Flavio and Kamal, Ojasv and Jin, Zhijing and Sch{\"o}lkopf, Bernhard},
  journal=arxiv,
  year={2023}
}

@article{taslp23_audiolm,
  title={{AudioLM}: a language modeling approach to audio generation},
  author={Borsos, Zal{\'a}n and Marinier, Rapha{\"e}l and Vincent, Damien and Kharitonov, Eugene and Pietquin, Olivier and Sharifi, Matt and Roblek, Dominik and Teboul, Olivier and Grangier, David and Tagliasacchi, Marco and others},
  journal=taslp,
  year={2023},
}

@inproceedings{icml23_audioldm,
  title={{AudioLDM}: Text-to-audio generation with latent diffusion models},
  author={Liu, Haohe and Chen, Zehua and Yuan, Yi and Mei, Xinhao and Liu, Xubo and Mandic, Danilo and Wang, Wenwu and Plumbley, Mark D},
  booktitle=icml,
  year={2023}
}

@article{arxiv23_audioldm2,
  title={{AudioLDM 2}: Learning holistic audio generation with self-supervised pretraining},
  author={Liu, Haohe and Tian, Qiao and Yuan, Yi and Liu, Xubo and Mei, Xinhao and Kong, Qiuqiang and Wang, Yuping and Wang, Wenwu and Wang, Yuxuan and Plumbley, Mark D},
  journal=arxiv,
  year={2023}
}

@inproceedings{nips23_melody,
  title={Efficient neural music generation},
  author={Lam, Max WY and Tian, Qiao and Li, Tang and Yin, Zongyu and Feng, Siyuan and Tu, Ming and Ji, Yuliang and Xia, Rui and Ma, Mingbo and Song, Xuchen and others},
  booktitle=nips,
  year={2023}
}

@article{arxiv23_audiobox,
  title={Audiobox: Unified Audio Generation with Natural Language Prompts},
  author={Vyas, Apoorv and Shi, Bowen and Le, Matthew and Tjandra, Andros and Wu, Yi-Chiao and Guo, Baishan and Zhang, Jiemin and Zhang, Xinyue and Adkins, Robert and Ngan, William and others},
  journal=arxiv,
  year={2023}
}

@article{arxiv23_musiclm,
  title={{MusicLM}: Generating music from text},
  author={Agostinelli, Andrea and Denk, Timo I and Borsos, Zal{\'a}n and Engel, Jesse and Verzetti, Mauro and Caillon, Antoine and Huang, Qingqing and Jansen, Aren and Roberts, Adam and Tagliasacchi, Marco and others},
  journal=arxiv,
  year={2023}
}

@inproceedings{cvpr_owens2016visually,
  title={Visually indicated sounds},
  author={Owens, Andrew and Isola, Phillip and McDermott, Josh and Torralba, Antonio and Adelson, Edward H and Freeman, William T},
  booktitle=cvpr,
  year={2016}
}

@inproceedings{eccvw_chen2018visually,
  title={Visually indicated sound generation by perceptually optimized classification},
  author={Chen, Kan and Zhang, Chuanxi and Fang, Chen and Wang, Zhaowen and Bui, Trung and Nevatia, Ram},
  booktitle=eccvw,
  year={2018}
}

@inproceedings{cvpr18_visual2sound,
  title={Visual to sound: Generating natural sound for videos in the wild},
  author={Zhou, Yipin and Wang, Zhaowen and Fang, Chen and Bui, Trung and Berg, Tamara L},
  booktitle=cvpr,
  year={2018}
}

@article{tip20_v2a,
  title={Generating visually aligned sound from videos},
  author={Chen, Peihao and Zhang, Yang and Tan, Mingkui and Xiao, Hongdong and Huang, Deng and Gan, Chuang},
  journal=tip,
  year={2020},
}

@inproceedings{nips23_diff_foley,
  title={{Diff-Foley}: Synchronized Video-to-Audio Synthesis with Latent Diffusion Models},
  author={Luo, Simian and Yan, Chuanhao and Hu, Chenxu and Zhao, Hang},
  booktitle=nips,
  year={2023}
}

@article{holm1979simple,
  title={A simple sequentially rejective multiple test procedure},
  author={Holm, Sture},
  journal={Scandinavian journal of statistics},
  year={1979},
  publisher={JSTOR}
}

@article{arxiv23_foleygen,
  title={{FoleyGen}: Visually-Guided Audio Generation},
  author={Mei, Xinhao and Nagaraja, Varun and Lan, Gael Le and Ni, Zhaoheng and Chang, Ernie and Shi, Yangyang and Chandra, Vikas},
  journal=arxiv,
  year={2023}
}

@inproceedings{casebeer2026generative,
  title={A generative-first neural audio autoencoder},
  author={Casebeer, Jonah and Zhu, Ge and Wang, Zhepei and Bryan, Nicholas J},
  booktitle=icassp,
  year={2026}
}

@article{bai2025dragon,
  title={{DRAGON}: Distributional Rewards Optimize Diffusion Generative Models},
  author={Bai, Yatong and Casebeer, Jonah and Sojoudi, Somayeh and Bryan, Nicholas J.},
  journal=tmlr,
  year={2025}
}

@inproceedings{novackpresto,
  title={Presto! Distilling Steps and Layers for Accelerating Music Generation},
  author={Novack, Zachary and Zhu, Ge and Casebeer, Jonah and McAuley, Julian and Berg-Kirkpatrick, Taylor and Bryan, Nicholas J.},
  booktitle=iclr,
  year={2025}
}

@article{team2025gemma,
  title={Gemma 3 technical report},
  author={Team, Gemma and Kamath, Aishwarya and Ferret, Johan and Pathak, Shreya and Vieillard, Nino and Merhej, Ramona and Perrin, Sarah and Matejovicova, Tatiana and Ram{\'e}, Alexandre and Rivi{\`e}re, Morgane and others},
  journal=arxiv,
  year={2025}
}

@inproceedings{peebles2023scalable,
  title={Scalable diffusion models with transformers},
  author={Peebles, William and Xie, Saining},
  booktitle=iccv,
  pages={4195--4205},
  year={2023}
}

@inproceedings{peeters2023self,
  title={Self-Similarity-Based and Novelty-Based Loss for Music Structure Analysis},
  author={Peeters, Geoffroy},
  booktitle={International Society of Music Information Retreival},
  year={2023}
}

@inproceedings{foote2001visualizing,
  title={Visualizing Musical Structure and Rhythm via Self-Similarity.},
  author={Foote, Jonathan and Cooper, Matthew},
  booktitle={ICMC},
  year={2001}
}

@inproceedings{shechtman2007matching,
  title={Matching local self-similarities across images and videos},
  author={Shechtman, Eli and Irani, Michal},
  booktitle=cvpr,
  year={2007},
}

@inproceedings{kwon2021learning,
  title={Learning self-similarity in space and time as generalized motion for video action recognition},
  author={Kwon, Heeseung and Kim, Manjin and Kwak, Suha and Cho, Minsu},
  booktitle=iccv,
  year={2021}
}

@inproceedings{bralios2025re,
  title={Re-Bottleneck: Latent Re-Structuring for Neural Audio Autoencoders},
  author={Bralios, Dimitrios and Casebeer, Jonah and Smaragdis, Paris},
  booktitle={MLSP},
  year={2025},
}

@article{bralios2025learning,
  title={Learning to Upsample and Upmix Audio in the Latent Domain},
  author={Bralios, Dimitrios and Smaragdis, Paris and Casebeer, Jonah},
  journal=arxiv,
  year={2025}
}

@article{sd_audio_ctrl,
  title={Cocola: Coherence-oriented contrastive learning of musical audio representations},
  author={Ciranni, Ruben and Mariani, Giorgio and Mancusi, Michele and Postolache, Emilian and Fabbro, Giorgio and Rodol{\`a}, Emanuele and Cosmo, Luca},
  journal=arxiv,
  year={2024}
}

@inproceedings{qi2025customized,
  title={Customized Condition Controllable Generation for Video Soundtrack},
  author={Qi, Fan and Ma, Kunsheng and Xu, Changsheng},
  booktitle=cvpr,
  year={2025}
}

@article{shan2025hunyuanvideo,
  title={Hunyuanvideo-foley: Multimodal diffusion with representation alignment for high-fidelity foley audio generation},
  author={Shan, Sizhe and Li, Qiulin and Cui, Yutao and Yang, Miles and Wang, Yuehai and Yang, Qun and Zhou, Jin and Zhong, Zhao},
  journal=arxiv,
  year={2025}
}

@inproceedings{ji2026diff,
  title={Diff-V2M: A Hierarchical Conditional Diffusion Model with Explicit Rhythmic Modeling for Video-to-Music Generation},
  author={Ji, Shulei and Wang, Zihao and Yu, Jiaxing and Yang, Xiangyuan and Li, Shuyu and Wu, Songruoyao and Zhang, Kejun},
  booktitle=aaai,
  year={2026}
}
